%% file: neurips_2026.tex
\documentclass{article}

 \usepackage[preprint]{neurips_2026}

\usepackage[utf8]{inputenc} 
\usepackage[T1]{fontenc}    
\usepackage{hyperref}       
\usepackage{url}            
\usepackage{booktabs}       
\usepackage{amsfonts}       
\usepackage{nicefrac}       
\usepackage{microtype}      
\usepackage{xcolor}         
\usepackage{amsmath}
\usepackage[pdftex]{graphicx}

\title{AdpSplit: Error-Driven Adaptive Splitting for Faster Geometry Discovery in 3D Gaussian Splatting}

\author{%
  Yongjae Lee \\
  Arizona State University\\
  Tempe, AZ, 85281 \\
  \texttt{ylee298@asu.edu} \\
  \And
  Jingxing Li \\
  Arizona State University\\
  Tempe, AZ, 85281 \\
  \texttt{jingxing@asu.edu} \\
  \And
  Abhay Kumar Yadav \\
  Johns Hopkins University\\
  Baltimore, MD, 21218 \\
  \texttt{ayadav13@jh.edu} \\
  \And
  Rama Chellappa \\
  Johns Hopkins University\\
  Baltimore, MD, 21218 \\
  \texttt{rchella4@jhu.edu} \\
  \And
  Deliang Fan\thanks{Corresponding author} \\
  Arizona State University\\
  Tempe, AZ, 85281 \\
  \texttt{dfan@asu.edu} \\
}

\begin{document}

\maketitle

\begin{figure}[h]
  \centering
  \includegraphics[width=\linewidth]{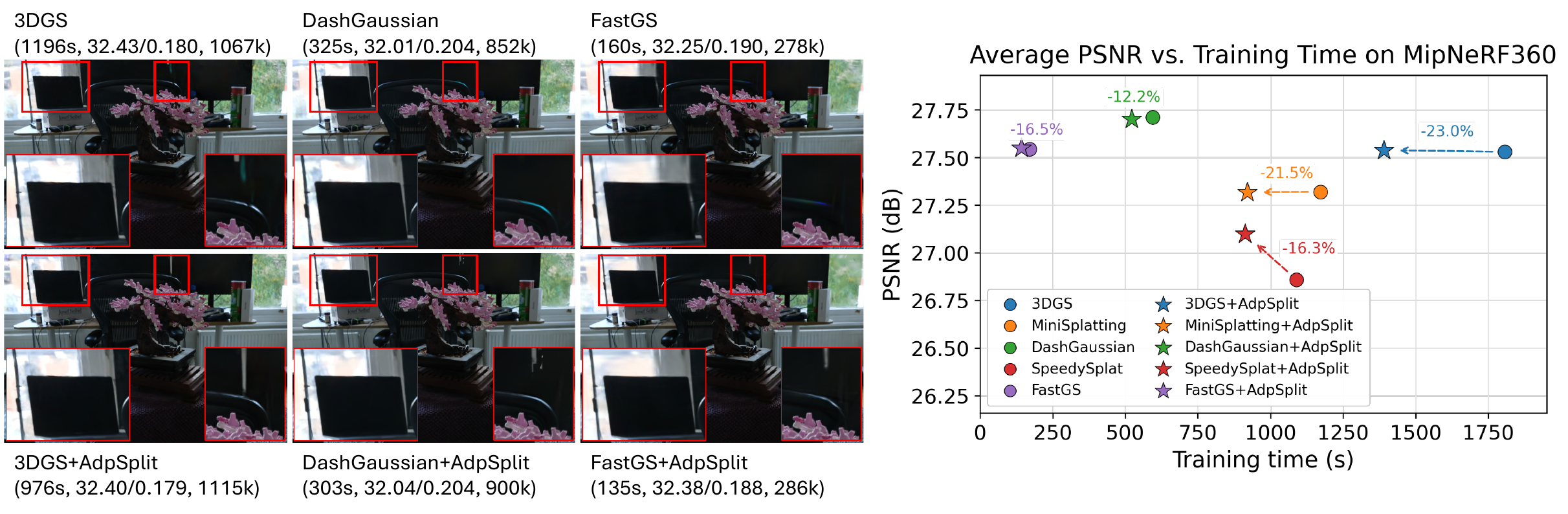}
  \caption{AdpSplit is a drop-in split operator for efficient 3D Gaussian
  Splatting (3DGS) training. Across multiple accelerated 3DGS baselines, its
  error-driven adaptive split improves reconstruction of under-modeled regions
  while reducing training time. Values reported for the bonsai scene are
  ordered as (training time, PSNR/LPIPS, number of Gaussians).
  Percentages in the graph indicate reductions in training time.}
  \label{fig:1}
\end{figure}

\begin{abstract}
Adaptive density control in 3D Gaussian Splatting (3DGS) repeatedly grows the Gaussian population through fixed-cardinality random splitting to discover useful scene structure. However, in vanilla 3DGS, its binary split operator requires many densification rounds to expose fine details, making it a bottleneck for efficient training schedules with fewer iterations. We introduce \textbf{AdpSplit}, an error-driven adaptive split operator that determines the number of split children and initializes the child parameters from L1-pixel-error region statistics, enabling fewer densification iterations, thus reduced training time, while preserving the rendering quality of full-schedule training. Across the MipNeRF360, Deep-Blending, and Tanks\&Temples datasets, AdpSplit reduces the training time of multiple accelerated 3DGS pipelines by 9.2\%--22.3\% as a simple drop-in replacement for the standard split operator. With FastGS, AdpSplit matches the full-schedule PSNR on MipNeRF360 while reducing training time by 16.4\%, corresponding to a 12.6$\times$ acceleration over vanilla 3DGS.
\end{abstract}

\section{Introduction}
Novel view synthesis (NVS) is a core computer vision task that aims to generate
photorealistic images from novel viewpoints given a set of observed images.
Neural radiance fields (NeRF)~\citep{Mildenhall2020NeRF} showed that a scene can
be represented as a continuous neural function, enabling high-fidelity view
synthesis. However, NeRF relies on dense volumetric sampling and repeated
forward passes through the neural networks, making both optimization and rendering
computationally expensive. In contrast, 3D Gaussian Splatting
(3DGS)~\citep{Kerbl20233DGaussian} introduced an explicit scene representation
based on learnable anisotropic Gaussian primitives and a differentiable rasterizer. 
With fast optimization, real-time rendering, and competitive visual quality, 3DGS has
become a practical foundation for downstream applications such as SLAM~\citep{Yan2024GSSLAM}, dynamic scene modeling~\citep{Wu20244DGaussian}, text-to-3D generation~\citep{Tang2023DreamGaussian}, and digital
humans~\citep{Qian2023GaussianAvatars}.

The quality and efficiency of 3DGS depend heavily
on how the Gaussian population is built during training. Starting from a sparse
structure-from-motion~\citep{Snavely2006Photo} point cloud,
the 3DGS pipeline repeatedly applies adaptive density control
(ADC), growing the Gaussian population to fill missing scene geometry.
In this sense, ADC is fundamentally a scene-structure search mechanism that
discovers a useful scene structure through repeated densification rounds.
Naive propagation, however, often requires many such rounds, thereby slowing
optimization and producing redundant Gaussians.

To accelerate optimization, one line of work reduces the number of Gaussians
through pruning, compression, or structured representations while leaving the standard ADC process
untouched~\citep{Girish2024EAGLES,Lee2024Compact3D,Zhang2024LP3DGS,Lu2023ScaffoldGS,Hanson2025Speedy}.
Another line of work addresses Gaussian propagation more directly through
elaborated split candidate selection, optimization-based splitting, or reinitializing
the scene~\citep{Fang2024MiniSplatting,Cheng2024GaussianPro,Wang2025SteepGS,Ren2025FastGS}.
These advances are important, and they mainly answer which Gaussians should be
split or removed, or where the propagated Gaussians should be placed. 
However, how many split children should be
produced for each split candidate, and what shapes those children should take,
remain unexplored.

\begin{figure}[t]
  \centering
  \includegraphics[width=\linewidth]{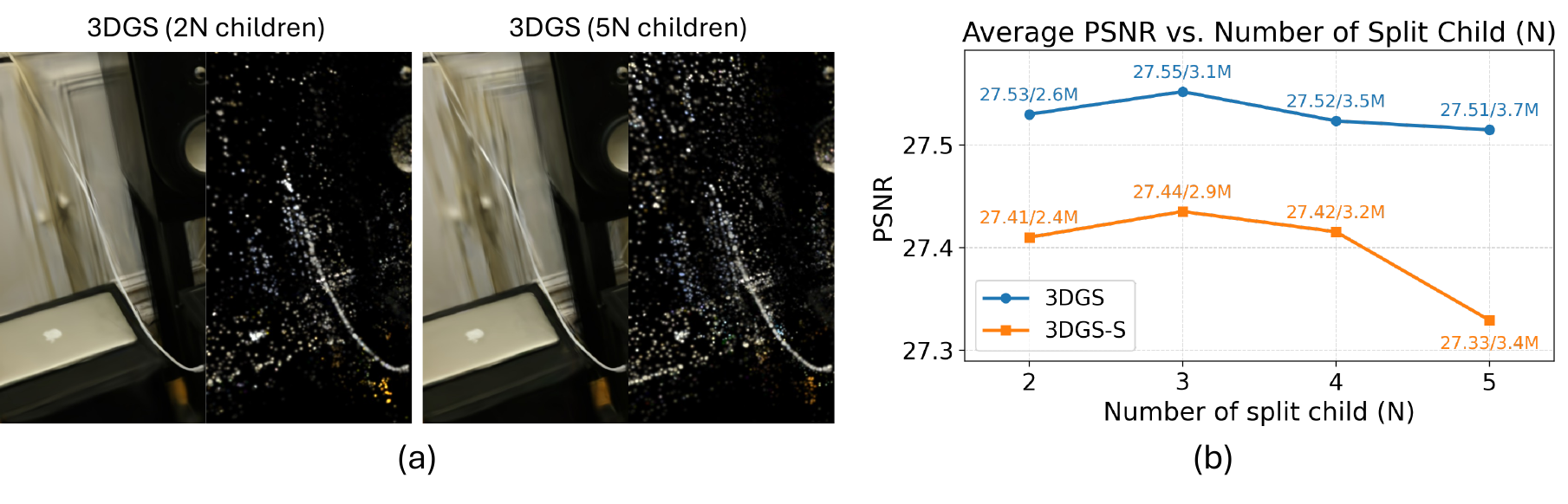}
  \caption{Using more randomly placed children can increase the risk of amplifying noise (a). Increasing the number of split children does not necessarily improve reconstruction quality (b). Values indicate PSNR/the number of Gaussians, and 3DGS-S is trained with the densification schedule shortened from 15k to 9k iterations.}
  \label{fig:2}
\end{figure}

The standard ADC split operator produces a fixed number of children, typically
two, for each split candidate and places them at randomly sampled locations.
This fixed-cardinality random splitting is a fundamental bottleneck. If a parent Gaussian
(i.e., a split candidate) covers a structure that eventually requires $G$ distinct Gaussians, a
binary split needs at least $\log_2 G$ densification rounds to expose that
structure.  A naive alternative is to split each parent
into more than two children, but additional children alone do not guarantee a
better scene representation. Randomly initialized children may redundantly fit
local observations, amplify noisy details, or miss the regions that actually
cause view loss. 
\autoref{fig:2}a visualizes this problem: 
the five-child split allocates additional Gaussians to
under-reconstructed regions, but places them poorly for reducing perceptual error.
This is especially problematic in shortened training schedules, where fewer
densification steps are available.
In our pilot study (\autoref{fig:2}b), 
a densification schedule shortened from 15k to 9k iterations with five children
per parent produces more Gaussians than a longer binary-split schedule, yet
achieves lower PSNR (27.33 PSNR with 3.4M Gaussians for 3DGS-S (5N) vs.
27.53 PSNR with 2.6M Gaussians for 3DGS (2N)). Thus, the core issue is not only how many Gaussians are added, but how those Gaussians are initialized.

We propose \textbf{AdpSplit}, an error-driven adaptive split method that replaces
fixed-cardinality random child placement with view-loss-aware child generation. 
During ADC, AdpSplit renders sampled training views, identifies high-error
pixels for which the candidate is a dominant contributor, and partitions these
pixels into connected error regions. Each region proposes one child, so the
number of children is determined by the observed error structure rather than a
global constant. AdpSplit then initializes each child by back-projecting the
region centroid to the depth that best agrees with the parent Gaussian,
estimating its covariance from the region's statistics, and initializing its
color from the ground-truth observation. Finally, cross-view child merging
removes redundant children generated from different views of the same portion of a Gaussian.
As shown in \autoref{fig:1}, AdpSplit can be used as a drop-in replacement
for the standard ADC split operator in existing 3DGS training accelerators, further
reducing training time while maintaining rendering quality.
Our contributions are summarized as follows:
\begin{itemize}
  \item We identify fixed-cardinality random splitting as a hidden bottleneck in
  fast 3DGS training and show that naively increasing the number of children can
  increase the Gaussian count while leaving reconstruction quality unchanged or
  even impairing it.
  \item We introduce AdpSplit, a new ADC split operator that adaptively determines
  the number of children, estimates their parameters from attributable
  L1-pixel-error region statistics, and merges redundant child proposals across views.
  \item We demonstrate that AdpSplit can be integrated with existing 3DGS
  acceleration pipelines, achieving comparable rendering quality to the baselines
  with a shortened training schedule.
\end{itemize}

\section{Related Works}
We review prior work on accelerating 3DGS training, focusing on scene compaction and adaptive density control.

\paragraph{Gaussian compaction.}
One common strategy is to reduce the Gaussian count. These methods learn
pruning masks, estimate uncertainty, or analyze importance statistics to prune
low-utility Gaussians that have little impact on reconstruction
quality~\citep{Fan2023LightGaussian,Zhang2024LP3DGS,Hanson2025PUP, Morgenstern2024Compact3D,Lee2024Compact3D,Niedermayr2024Compressed,Niemeyer2024RadSplat,Hanson2025Speedy}.
Other methods focus on parameter-level information compression via entropy
encoding or quantization, or on Gaussian-level compression
through structured representations~\citep{Girish2024EAGLES,Lu2023ScaffoldGS,Ren2025OctreeGS,Chen2025HAC,Chen2025HAC++}.
These approaches are complementary to AdpSplit. They mainly compress, encode, or
sparsify the representation after Gaussians have been created, whereas our
method targets the earlier scene-building decision of how split children are
generated.

\paragraph{Adaptive density control.}
Adaptive density control~\citep{Kerbl20233DGaussian} grows the 3DGS scene representation from sparse initialization by splitting and cloning Gaussians selected using image-space gradient statistics. 
Later work has shown that
this heuristic can be inefficient and unstable, motivating more principled
growth rules, budgeted selection, depth reinitialization, optimization-based splitting, 
and error-guided insertion~\citep{BulO2024Revising,Fang2024MiniSplatting,Cheng2024GaussianPro,Kheradmand20243DGSMCMC,Wang2025SteepGS,Ren2025FastGS,Chen2025DashGaussian,Baranowski2025ConeGS,Gao2025EasySplat}.
These works reduce redundant growth, but the standard split operation commonly
remains fixed-cardinality, and the initialization of new Gaussians is not
directly integrated with the splitting and cloning operations.


\paragraph{Positioning of AdpSplit.}
To the best of our knowledge, no existing work has shown that an effective split
operation can expedite the discovery of useful scene structure.
The Gaussian initialization strategies in MiniSplatting~\citep{Fang2024MiniSplatting}
and ConeGS~\citep{Baranowski2025ConeGS} can improve the spatial distribution of
Gaussians, but they can substantially increase the number of Gaussians that must
be optimized or require external geometric priors. 
FastGS~\citep{Ren2025FastGS} and DashGaussian~\citep{Chen2025DashGaussian}
are designed for fast schedules, but
they primarily regulate candidate selection or Gaussian budgets 
rather than estimating the number and parameters of split children
from local error structure. 
SteepGS~\citep{Wang2025SteepGS} is particularly relevant because it asks which
direction the children should be deployed; however, it does not consider how
many children to create or where exactly they should be placed.
Our approach is orthogonal to existing works, as supported by the
experiments in \autoref{sec:experiments}.

\section{Preliminaries: 3D Gaussian Splatting}
\label{ssec:prelim}

3DGS represents a scene as a set of $G$ Gaussians
$\mathcal{G} = \{\mathcal{G}_i\}_{i=1}^{G}$, each parameterized by a mean
$\boldsymbol{\mu}_i \in \mathbb{R}^3$, a 3D covariance
$\boldsymbol{\Sigma}_i = \mathbf{R}_i \mathbf{S}_i \mathbf{S}_i^\top \mathbf{R}_i^\top$
with rotation matrix $\mathbf{R}_i$ and diagonal scale matrix
$\mathbf{S}_i=\operatorname{diag}(\mathbf{s}_i)$ with
$\mathbf{s}_i \in \mathbb{R}^3$, an opacity
$o_i \in (0,1)$, and spherical harmonic (SH) color coefficients
$\mathbf{c}_i=\{\mathbf{c}_i^{\mathrm{dc}},\mathbf{c}_i^{\mathrm{rest}}\}$,
consisting of a DC component and higher-order coefficients.
Gaussians are projected to the image plane and $\alpha$-blended in depth order
with a differentiable rasterizer. For a pixel $u$ in view $v$, the rendered
color is
\begin{equation}
  \hat{\mathbf{I}}_v(u) =
  \sum_{i \in \mathcal{O}_v(u)} T_{i,v}(u)\,\alpha_{i,v}(u)\,\mathbf{c}_i(\mathbf{d}_v),
  \quad
  T_{i,v}(u)=\prod_{j<i}\bigl(1-\alpha_{j,v}(u)\bigr),
  \label{eq:rendering}
\end{equation}
where $\mathcal{O}_v(u)$ is the set of Gaussians overlapping pixel $u$ in view
$v$ in front-to-back depth order, $\alpha_{i,v}(u)$ combines the learned opacity
with the projected 2D Gaussian footprint, $T_{i,v}(u)$ is the accumulated
transmittance, and $\mathbf{c}_i(\mathbf{d}_v)$ is the view-dependent color
decoded from the SH coefficients along viewing direction $\mathbf{d}_v$.

The ADC accumulates image-space gradients for visible Gaussians.
At each densification interval, it first computes the average gradient statistic
$g_i$ for each Gaussian. Among high-gradient Gaussians,
those with small spatial scale are cloned, while those with large spatial scale
are split. The split candidate set $\mathcal{S}$ is defined as
\begin{equation}
  i \in \mathcal{S}
  \quad\Longleftrightarrow\quad
  g_i \geq \tau_g
  \;\land\;
  \max(\mathbf{s}_i) > \tau_s,
  \label{eq:split_candidate}
\end{equation}
where $\tau_g$ is the densification gradient threshold and $\tau_s$ is the scale
threshold.

For each selected parent $\mathcal{G}_i$, the vanilla split operator generates a
fixed number $N$ of children, with $N=2$ by default. Each child $n$ is placed
near the parent by sampling an offset from the parent's Gaussian distribution:
\begin{equation}
  \boldsymbol{\delta}_{i,n} \sim
  \mathcal{N}(\mathbf{0}, \operatorname{diag}(\mathbf{s}_i^2)), \quad
  \boldsymbol{\mu}_{i,n}' =
  \boldsymbol{\mu}_i + \mathbf{R}_i \boldsymbol{\delta}_{i,n}.
  \label{eq:vanilla_split_sample}
\end{equation}
The remaining child parameters are initialized by copying the parent's rotation,
opacity, and SH coefficients, while uniformly shrinking the parent scale with
shrink factor $\eta$:
\begin{equation}
  \mathbf{s}_{i,n}' = \frac{\mathbf{s}_i}{\eta N}, \quad
  \mathbf{R}_{i,n}' = \mathbf{R}_i, \quad
  o_{i,n}' = o_i, \quad
  \mathbf{c}_{i,n}' = \mathbf{c}_i.
  \label{eq:vanilla_split_scale}
\end{equation}
After appending the children to the Gaussian set, the selected parents are
removed.

\section{Method}
\label{sec:method}

\begin{figure}[t]
  \centering
  \includegraphics[width=\linewidth]{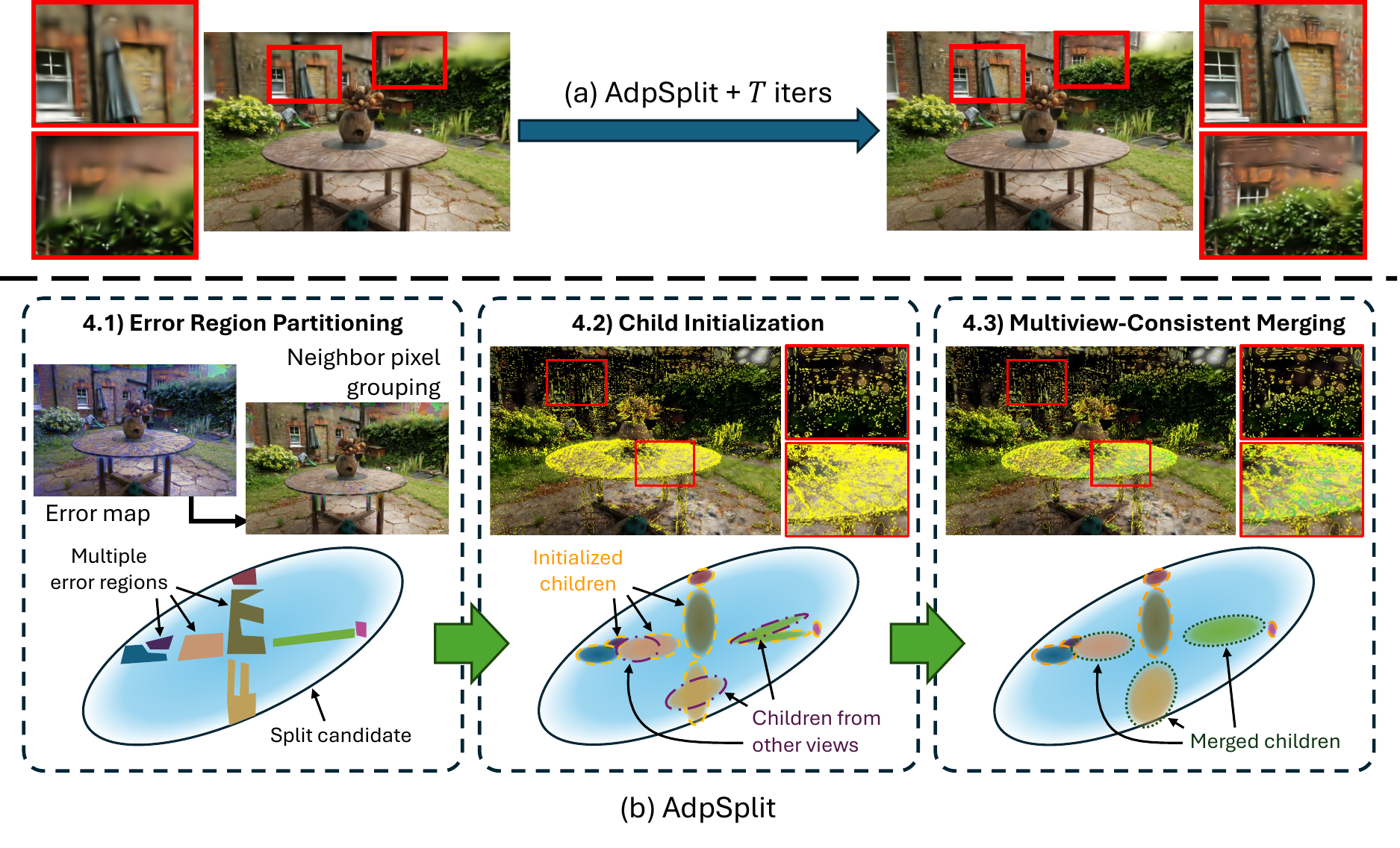}
  \caption{Overview of AdpSplit. Our method generates more children for a
  Gaussian when scattered error regions are detected, 
  enabling fine scene structures to be discovered in
  earlier training iterations (a). Neighboring high-error pixels attributed to
  the same split candidate are grouped as an error region, whose statistics
  initialize a child Gaussian. To eliminate redundant children from
  the same error portion, children from different views with similar positions
  and colors are merged (b).}
  \label{fig:3}
\end{figure}

\autoref{fig:3} gives a high-level overview of AdpSplit. At each densification
step, we use the standard 3DGS gradient-and-size rule
(\autoref{eq:split_candidate}) to obtain the split candidate set
$\mathcal{S}$, while small high-gradient Gaussians are still handled by the
original clone operation. For each candidate in $\mathcal{S}$, AdpSplit proceeds
in three stages: \textbf{(1) error region partitioning}, which attributes
high-error pixels to the candidate and groups them into error regions
(\autoref{ssec:erp}); \textbf{(2) child initialization}, which initializes one
or more children from the statistics of the error regions (\autoref{ssec:child_init}); and
\textbf{(3) cross-view child merging}, which removes redundant children 
across views before insertion to the Gaussian set (\autoref{ssec:merge}).

\subsection{Error Region Partitioning}
\label{ssec:erp}


We randomly sample $V$ training cameras and render them. During rendering, we
also collect a dominant Gaussian map $D_v$ by selecting, for each pixel, the
Gaussian with the largest alpha-weighted contribution:
\begin{equation}
  D_v(u) = \arg\max_{i \in \mathcal{O}_v(u)}
  T_{i,v}(u)\,\alpha_{i,v}(u),
  \label{eq:dominant-map}
\end{equation}
Then, we compute a min-max-normalized per-pixel L1 error map:
\begin{equation}
  \mathcal{E}_v(u) = \frac{|\hat{\mathbf{I}}_v(u) - \mathbf{I}_v^{\mathrm{gt}}(u)|_1
                    - \mathcal{E}_{\min}}{\mathcal{E}_{\max} - \mathcal{E}_{\min}}.
  \label{eq:l1map}
\end{equation}
where $\mathcal{E}_{\min}$ and $\mathcal{E}_{\max}$ denote the minimum and
maximum values of $|\hat{\mathbf{I}}_v(u)-\mathbf{I}_v^{\mathrm{gt}}(u)|_1$
over all pixels in view $v$.
A binary metric map $\mathcal{M}_v(u) = \mathbf{1}[\mathcal{E}_v(u) > \tau_{\mathrm{l1}}]$
marks high-error pixels, where $\tau_{\mathrm{l1}}$ is the normalized L1-error
threshold. Optionally, an erosion
filter with length $r_{\mathrm{erode}}$
removes isolated noisy pixels before error-region
grouping.

We stratify the normalized error values into $L$ error-level bands over
$[\tau_{\mathrm{l1}},1]$, and denote the resulting band index by $B_v(u)$.
For every split candidate $\mathcal{G}_i$ with $i \in \mathcal{S}$ and sampled
view $v$, distinct error regions are defined as maximal groups of neighboring
high-error pixels that are dominated by the same candidate and share the same
error band. Formally, the $q$-th error region for candidate $i$ in view $v$ is
\begin{equation}
  \mathcal{R}_v^q
  =
  \left\{
    u
    \;\middle|\;
    \mathcal{M}_v(u)=1,\;
    D_v(u)=i,\;
    u \sim_8 u^q,\;
    B_v(u)=B_v(u^q)
  \right\},
  \label{eq:error-region}
\end{equation}
where $u \sim_8 u^q$ means that $u$ can be reached from the seed pixel $u^q$
through a chain of 8-neighbor pixels satisfying the same conditions, and
$u^q$ satisfies $\mathcal{M}_v(u^q)=1$ and $D_v(u^q)=i$.
The region area
is defined as $\mathcal{A}_v^q=|\mathcal{R}_v^q|$, and only regions satisfying
\begin{equation}
  \mathcal{A}_v^q \ge m_{\min}
  \label{eq:min_region_area}
\end{equation}
are retained.

For every retained region $\mathcal{R}_v^q$ of candidate $\mathcal{G}_i$, we
compute region statistics: centroid $(\bar{x}_v^q,\bar{y}_v^q)$, area
$\mathcal{A}_v^q$, dominant PCA direction
$\mathbf{e}_{1,v}^q=(e_{1x,v}^q,e_{1y,v}^q)$, orthogonal direction
$\mathbf{e}_{2,v}^q=(-e_{1y,v}^q,e_{1x,v}^q)$, and standard deviations
$(\sigma_{1,v}^q,\sigma_{2,v}^q)$ from the pixel covariance,
as well as the ground-truth RGB value $\mathbf{r}_v^q$ at the centroid.  These statistics
are the sufficient information used to initialize 3D children.

\subsection{Child Initialization}
\label{ssec:child_init}

\begin{figure}[t]
    \centering
    \includegraphics[width=.7\linewidth]{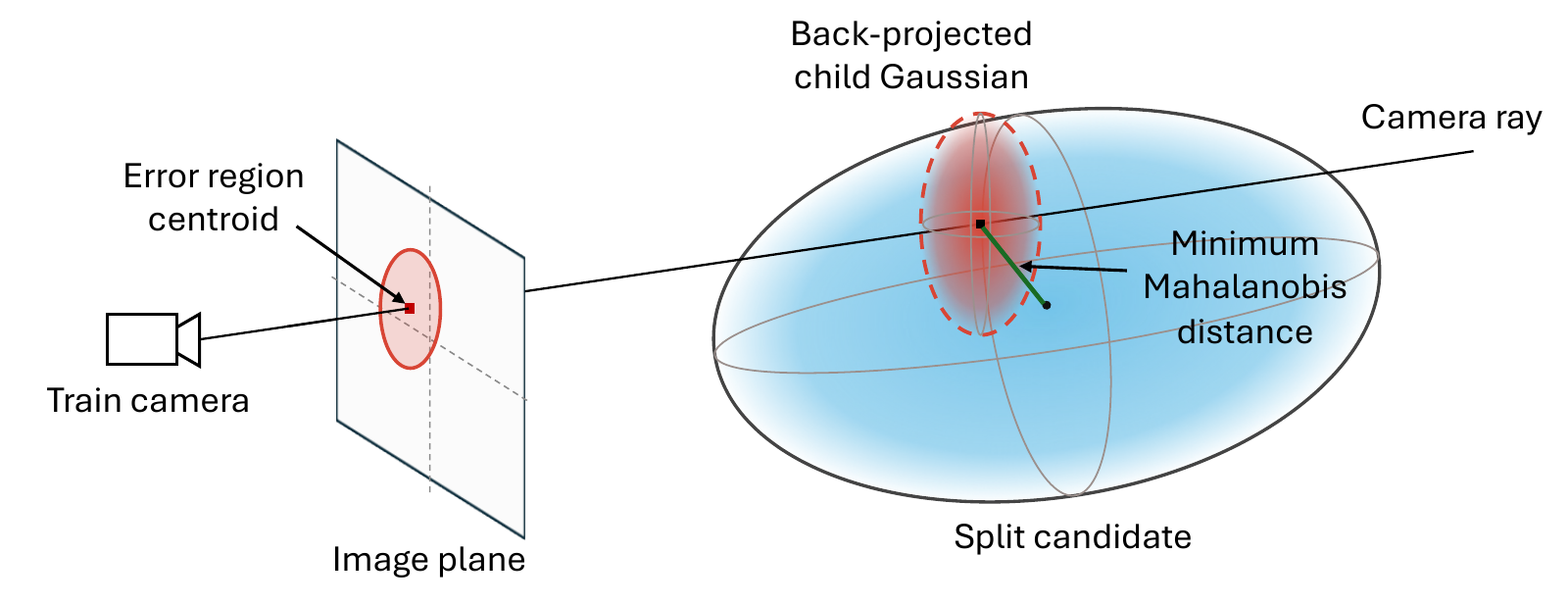}
    \caption{Child initialization from an attributed 2D error region. The
    region centroid defines a camera ray, and the child center is placed at the
    closest point to the parent center in the parent's Mahalanobis metric. The
    region covariance determines the child's rotations and scales.}
    \label{fig:4}
\end{figure}

Each attributed error region proposes one child Gaussian. The child is initialized
from the region's image-space position, shape, and color, while the parent
supplies a depth prior.

To compute the child center $\boldsymbol{\mu}_v^q$, we back-project the centroid
$(\bar{x}_v^q,\bar{y}_v^q)$ along its camera ray
\begin{equation}
  \boldsymbol{\ell}_v^q(t) = \mathbf{o}_v + t\hat{\mathbf{d}}_v^q,
  \label{eq:ray}
\end{equation}
where $\mathbf{o}_v$ is the camera center and $\hat{\mathbf{d}}_v^q$ is the
unit ray direction in world coordinates. We place the child at the point on this
ray that minimizes the Mahalanobis distance to the parent center, so that the
placement respects the parent's anisotropic Gaussian shape:
\begin{equation}
  t_v^* = \arg\min_t
  \left(\boldsymbol{\ell}_v^q(t)-\boldsymbol{\mu}_i\right)^\top
  \boldsymbol{\Sigma}_i^{-1}
  \left(\boldsymbol{\ell}_v^q(t)-\boldsymbol{\mu}_i\right)
  =
  \frac{(\boldsymbol{\mu}_i - \mathbf{o}_v)^\top \boldsymbol{\Sigma}_i^{-1}
        \hat{\mathbf{d}}_v^q}
       {(\hat{\mathbf{d}}_v^q)^\top \boldsymbol{\Sigma}_i^{-1} \hat{\mathbf{d}}_v^q + \epsilon},
  \label{eq:tstar}
\end{equation}
where $\epsilon$ is a small constant for numerical stability (\autoref{fig:4}). 
We then initialize
the child parameters as
\begin{equation}
  \begin{aligned}
    \boldsymbol{\mu}_v^q &= \boldsymbol{\ell}_v^q(t_v^*), \quad
    \mathbf{R}_v^q =
    [\hat{\mathbf{u}}_1 \mid \hat{\mathbf{u}}_2 \mid \mathbf{z}_v], \\
    \mathbf{s}_v^q
    &= \left(\|\mathbf{a}_{1,v}^q\|,\|\mathbf{a}_{2,v}^q\|,
       \|\mathbf{a}_{2,v}^q\|\right), \quad
    \mathbf{c}_v^q = \mathbf{r}_v^q, \quad
    o_v^q = o_i.
  \end{aligned}
  \label{eq:child_params}
\end{equation}
where $\hat{\mathbf{u}}_1$ and $\hat{\mathbf{u}}_2$ are orthonormal in-plane
axes obtained from the unprojected PCA axes $\mathbf{a}_{1,v}^q$ and
$\mathbf{a}_{2,v}^q$, $\mathbf{z}_v$ is the camera forward direction, and
the opacity is inherited from the parent. To conservatively initialize the
child, we set its thickness along the camera forward direction to the minimum
scale in image-view space, i.e., $\|\mathbf{a}_{2,v}^q\|$.
\autoref{sec:child_position_derivation} gives the full initialization details.

\subsection{Cross-View Child Merging}
\label{ssec:merge}

The same underfit portion of a Gaussian can be visible in several sampled views, so the
previous stage may produce redundant child proposals for one parent.  We merge
compatible proposals before inserting them into the Gaussian set $\mathcal{G}$.

For each parent independently, two child proposals $\mathcal{G}^{(i)}$ and
$\mathcal{G}^{(j)}$ are connected if their symmetric Mahalanobis distance is
below the distance threshold $\gamma_d$ and their maximum RGB-channel difference
is below the color threshold $\gamma_c$:
\begin{equation}
  \sqrt{\boldsymbol{\delta}^\top \boldsymbol{\Sigma}_i^{-1} \boldsymbol{\delta}}
  + \sqrt{\boldsymbol{\delta}^\top \boldsymbol{\Sigma}_j^{-1} \boldsymbol{\delta}}
  \leq \gamma_d
  \quad\text{and}\quad
  \|\mathbf{c}^{(i)}-\mathbf{c}^{(j)}\|_\infty \leq \gamma_c,
  \label{eq:merge_conditions}
\end{equation}
where $\boldsymbol{\delta} = \boldsymbol{\mu}^{(j)} - \boldsymbol{\mu}^{(i)}$.

We group connected proposals in the merge graph, so merge eligibility is
transitive. For a merged group $\mathcal{M}$ with $M=|\mathcal{M}|$ proposals,
the center and color are the member averages.  The covariance is chosen to
enclose all member ellipsoids.  Let
$\mathbf{E}$ be the eigenvectors of the mean covariance
$\bar{\boldsymbol{\Sigma}}_{\mathcal{M}}=\frac{1}{M}\sum_{m \in \mathcal{M}}\boldsymbol{\Sigma}^{(m)}$.  Along
each axis $\mathbf{e}_r$ of $\mathbf{E}$,
we set the merged variance by the maximum member reach:
\begin{equation}
  \begin{aligned}
    \boldsymbol{\mu}_{\mathcal{M}}
    &= \frac{1}{M}\sum_{m \in \mathcal{M}}\boldsymbol{\mu}^{(m)}, \quad
    \boldsymbol{\Sigma}_{\mathcal{M}}
    = \mathbf{E}\,\operatorname{diag}(\boldsymbol{\lambda}_{\mathcal{M}})\,\mathbf{E}^\top, \quad
    \mathbf{c}_{\mathcal{M}}
    = \frac{1}{M}\sum_{m \in \mathcal{M}}\mathbf{c}^{(m)}, \\
    \lambda_{\mathcal{M},r}
    &= \max_{m \in \mathcal{M}}\!\bigl(
      |\mathbf{e}_r^\top(\boldsymbol{\mu}^{(m)} - \boldsymbol{\mu}_{\mathcal{M}})|
      + \sigma_{m,r}\bigr)^2,\quad r \in \{1,2,3\}.
  \end{aligned}
  \label{eq:merge_cov}
\end{equation}
where $\sigma_{m,r}$ is the $1\sigma$ extent of member $m$ along $\mathbf{e}_r$.

After merging, each split candidate has a final set of child proposals. 
To avoid letting a single parent introduce excessive children, we sort its children by
covariance extent and keep the largest $N_{\max}$ proposals.


\section{Experiments}
\label{sec:experiments}

\subsection{Setup}
In our experiments, we train 3DGS~\citep{Kerbl20233DGaussian} and various 3DGS training accelerators with AdpSplit and compare their train time and reconstruction quality.

\paragraph{Metrics.}
We report novel view synthesis quality using PSNR (dB), SSIM, and LPIPS on the test views.
To measure training efficiency, we also report wall-clock training time in
seconds, rendering FPS, and the final number of Gaussians.

\paragraph{Datasets.}
We evaluate on the standard real-world benchmarks: MipNeRF360~\citep{Barron2022Mipnerf360}, 
Tanks\&Temples~\citep{Knapitsch2017Tanks}, and
Deep-Blending~\citep{Hedman2018Deep}. Following the 3DGS protocol~\citep{Kerbl20233DGaussian}, images are resized and split into training and
test views.

\paragraph{Baselines.}
We select 3DGS~\citep{Kerbl20233DGaussian},
MiniSplatting~\citep{Fang2024MiniSplatting},
DashGaussian~\citep{Chen2025DashGaussian}, SpeedySplat~\citep{Hanson2025Speedy}, and
FastGS~\citep{Ren2025FastGS} as our baselines. We train each using its original
training schedule. We also train each with a shortened densification schedule, denoted by \emph{-S}
(densification stops at 9k iterations, followed by continued optimization until 24k iterations).
We apply AdpSplit on top of the shortened densification variants, and denote the
variants by \emph{-AdpSplit}. For DashGaussian's shortened schedule, we use 19k
densification iterations and 5k fine-tuning iterations, and use the same \emph{-S}
and \emph{-AdpSplit} suffixes.

\paragraph{Implementation.}
AdpSplit is invoked only inside the densification step and leaves the baseline
rendering loss and optimizer unchanged. Following the baselines' setting, the
densification interval is $T=100$ for 3DGS, MiniSplatting, DashGaussian, and
SpeedySplat, and $T=500$ for FastGS. For
MiniSplatting-S and MiniSplatting-AdpSplit, we use \texttt{reinit\_interval 3000},
\texttt{simp\_iteration1 9000}, and \texttt{simp\_iteration2 14000}
to adapt the original setting to the shortened densification schedule. 
For the same reason, we use \texttt{densify\_until\_iter 9000} for SpeedySplat-S and SpeedySplat-AdpSplit. 

For AdpSplit, we set the L1-error threshold $\tau_{\mathrm{l1}}=0.1$,
erosion length $r_{\mathrm{erode}}=2$, minimum error-region area
$m_{\min}=5$, number of error-level bands $L=3$, maximum children per parent
$N_{\max}=19$, and number of sampled views $V=20$ by default. For outdoor
scenes from MipNeRF360 and the \texttt{playroom} scene from Deep-Blending, we reduce the
child cap and sampled-view count to $N_{\max}=9$ and $V=10$. 
When merging cross-view children, we use $\gamma_d=2.0$ and $\gamma_c=0.15$.
The split thresholds $\tau_s$ and $\tau_g$, along with all other hyper-parameters, follow
the corresponding baseline defaults.

Our method generates an adaptive set of $N_i$ split children with proper initial parameter
estimates, according to the available error evidence for split candidates. 
When appending the estimated children to the Gaussian set,
we additionally include a parent copy with opacity reduced by a factor of $N_i+1$ and scale unchanged
to cover error-perfect regions and keep the total opacity unchanged
before and after the split operation.
If no sampled view attributes error
pixels to the candidate (when the Gaussian has never been dominant in the sampled views), 
the candidate falls back to the vanilla split with $N=2$ children. 
When error pixels are attributed
but no valid error region survives 
the pixel erosion filter and minimum error-region area filtering,
we reset the optimization status of the parent.

We modified the CUDA rasterizer of 3DGS~\citep{Kerbl20233DGaussian}
to use the compact-box method~\citep{Ren2025FastGS} 
for rasterizing Gaussians and to additionally 
output dominant Gaussian maps for sampled views.
We implemented a CUDA extension for projecting split candidates and collecting
high-error pixels inside each candidate footprint, and utilized the
\texttt{cuCIM} library to group neighboring error pixels into error regions.
The rest of AdpSplit is implemented in PyTorch.

\subsection{Results}
In this section, we empirically demonstrate the effectiveness of AdpSplit
by comparing baselines and AdpSplit on top of the baselines, and the effect of each stage by progressively ablating them. All experiments are conducted using NVIDIA RTX 5000 Ada. For more visualization and sensitivity of hyperparameters, please refer to~\autoref{sec:additional_results} in Appendix.

\begin{table}[t]
  \caption{Average reconstruction quality and efficiency on the three benchmarks. Time denotes training time in seconds, $\Delta t$ denotes the percentage change in training time from the baseline of each triple, and \#G. denotes the number of Gaussians in millions. Bold indicates the best value within each baseline triple.}
  \label{tab:1}
  \centering
  \setlength{\tabcolsep}{2pt}
  \resizebox{\linewidth}{!}{%
  \begin{tabular}{@{}lrrrrrrrrrrrrrrrrrr@{}}
    \toprule
    \textbf{Method} & \multicolumn{6}{c}{\textbf{MipNeRF360}} & \multicolumn{6}{c}{\textbf{Deep-Blending}} & \multicolumn{6}{c}{\textbf{Tanks\&Temples}} \\
    \cmidrule(lr){2-7} \cmidrule(lr){8-13} \cmidrule(lr){14-19}
      & \textbf{PSNR} & \textbf{SSIM} & \textbf{LPIPS} & \textbf{Time} & \textbf{$\Delta t$} & \textbf{\#G.} & \textbf{PSNR} & \textbf{SSIM} & \textbf{LPIPS} & \textbf{Time} & \textbf{$\Delta t$} & \textbf{\#G.} & \textbf{PSNR} & \textbf{SSIM} & \textbf{LPIPS} & \textbf{Time} & \textbf{$\Delta t$} & \textbf{\#G.} \\
    \midrule
    3DGS & 27.53 & \textbf{0.813} & \textbf{0.221} & 1807 & -- & 2.65 & 29.68 & \textbf{0.907} & \textbf{0.239} & 1676 & -- & 2.48 & 23.72 & \textbf{0.852} & \textbf{0.169} & 967 & -- & 1.58 \\
    3DGS-S & 27.41 & 0.808 & 0.228 & \textbf{1318} & \textbf{-27.1} & \textbf{2.41} & \textbf{29.87} & 0.906 & 0.243 & \textbf{1152} & \textbf{-31.3} & \textbf{2.21} & 23.69 & 0.850 & 0.174 & \textbf{707} & \textbf{-26.9} & \textbf{1.47} \\
    3DGS-AdpSplit & \textbf{27.54} & 0.810 & 0.225 & 1391 & -23.0 & 2.55 & 29.80 & \textbf{0.907} & 0.240 & 1197 & -28.6 & 2.25 & \textbf{23.83} & 0.850 & 0.172 & 748 & -22.6& 1.51 \\
    \midrule
    MiniSplatting & \textbf{27.32} & \textbf{0.822} & \textbf{0.217} & 1172 & -- & 0.49 & 29.97 & 0.910 & \textbf{0.241} & 1050 & -- & 0.55 & \textbf{23.40} & \textbf{0.846} & \textbf{0.181} & 723 & -- & 0.30 \\
    MiniSplatting-S & 27.27 & 0.821 & 0.219 & \textbf{898} & \textbf{-23.4} & \textbf{0.48} & \textbf{30.07} & \textbf{0.911} & 0.242 & \textbf{816} & \textbf{-22.3}& \textbf{0.53} & \textbf{23.40} & 0.843 & 0.186 & \textbf{527} & \textbf{-27.1}& \textbf{0.28} \\
    MiniSplatting-AdpSplit & \textbf{27.32} & 0.820 & 0.220 & 920 & -21.5 & \textbf{0.48} & 29.97 & 0.910 & 0.242 & 828 & -21.1 & \textbf{0.53} & 23.27 & 0.838 & 0.192 & 562 & -22.3& \textbf{0.28} \\
    \midrule
    DashGaussian & \textbf{27.71} & \textbf{0.820} & \textbf{0.213} & 594 & -- & 2.43 & \textbf{29.79} & \textbf{0.905} & \textbf{0.246} & 358 & -- & 1.95 & 24.00 & \textbf{0.853} & \textbf{0.178} & 349 & -- & 1.21 \\
    DashGaussian-S & 27.63 & 0.819 & 0.214 & \textbf{499} & \textbf{-16.0}& 2.40 & 29.72 & 0.904 & 0.248 & \textbf{314} & \textbf{-12.3}& \textbf{1.82} & 23.94 & 0.850 & 0.181 & \textbf{292} & \textbf{-16.3}& 1.17 \\
    DashGaussian-AdpSplit & 27.70 & \textbf{0.820} & \textbf{0.213} & 522 & -12.1& \textbf{2.36} & 29.71 & 0.904 & 0.248 & 325 & -9.2& 1.85 & \textbf{24.12} & 0.851 & 0.179 & 304 & -12.9 & \textbf{1.15} \\
    \midrule
    SpeedySplat & 26.86 & 0.781 & 0.295 & 1090 & -- & \textbf{0.30} & 29.63 & 0.903 & 0.268 & 841 & -- & \textbf{0.25} & \textbf{23.51} & 0.820 & 0.240 & 500 & -- & \textbf{0.18} \\
    SpeedySplat-S & 27.03 & 0.787 & 0.283 & \textbf{868} & \textbf{-20.4}& 0.34 & \textbf{29.64} & \textbf{0.904} & \textbf{0.266} & \textbf{649} & \textbf{-22.8}& 0.27 & 23.46 & 0.823 & 0.234 & \textbf{395} & \textbf{-20.9} & 0.20 \\
    SpeedySplat-AdpSplit & \textbf{27.10} & \textbf{0.790} & \textbf{0.279} & 912 & -16.3 & 0.35 & 29.62 & 0.903 & \textbf{0.266} & 677 & -19.5& 0.27 & 23.49 & \textbf{0.824} & \textbf{0.233} & 422 & -15.5 & 0.20 \\
    \midrule
    FastGS & 27.54 & \textbf{0.798} & 0.261 & 171 & -- & 0.40 & 29.90 & \textbf{0.905} & \textbf{0.267} & 105 & -- & 0.22 & \textbf{24.26} & \textbf{0.842} & \textbf{0.208} & 108 & -- & 0.24 \\
    FastGS-S & 27.37 & 0.790 & 0.272 & \textbf{127} & \textbf{-25.7}& \textbf{0.31} & 29.80 & 0.901 & 0.274 & \textbf{78} & \textbf{-25.7}& \textbf{0.17} & 23.99 & 0.837 & 0.214 & \textbf{80} & \textbf{-25.9} & \textbf{0.21} \\
    FastGS-AdpSplit & \textbf{27.55} & 0.797 & \textbf{0.260} & 143 & -16.4& 0.40 & \textbf{29.95} & 0.904 & 0.269 & 83 & -21.0& 0.19 & 24.17 & 0.840 & 0.209 & 86 & -20.4& 0.23 \\
    \bottomrule
  \end{tabular}%
  }
\end{table}

\subsubsection{Improving existing efficient training methods}
\autoref{tab:1} shows that shortened densification schedules
consistently reduce training time, but often at the cost of reconstruction
quality. This trade-off is especially visible for accelerated training
pipelines. On MipNeRF360, FastGS-S reduces training time by 25.7\% but also
decreases PSNR from 27.54 to 27.37. Similar quality drops appear for FastGS on
Deep-Blending and Tanks\&Temples, indicating that simply removing densification
rounds can leave scene structure under-modeled even when split candidates are
carefully selected.

AdpSplit recovers much of this lost quality while preserving most of the speedup
from the shortened schedules. For 3DGS on MipNeRF360, the shortened schedule
reduces PSNR by 0.12 dB, whereas 3DGS-AdpSplit restores the PSNR to 27.54
while still reducing training time by 23.0\% relative to the full 3DGS baseline.
The same trend holds for FastGS: FastGS-AdpSplit matches the full FastGS
PSNR on MipNeRF360, improves LPIPS from 0.261 to 0.260, and trains in only
143 seconds, corresponding to a 16.4\% time reduction. Across Deep-Blending,
AdpSplit maintains comparable SSIM for most baselines while reducing training
time by up to 28.6\%. Remarkably, FastGS-AdpSplit trains a scene 
from Deep-Blending in only 83 seconds. 
On Tanks\&Temples, AdpSplit similarly keeps quality metrics close to the full
baselines while reducing training time by 12.9\% to 22.6\% depending on the
underlying method.

\begin{figure}[t]
    \centering
    \includegraphics[width=\linewidth]{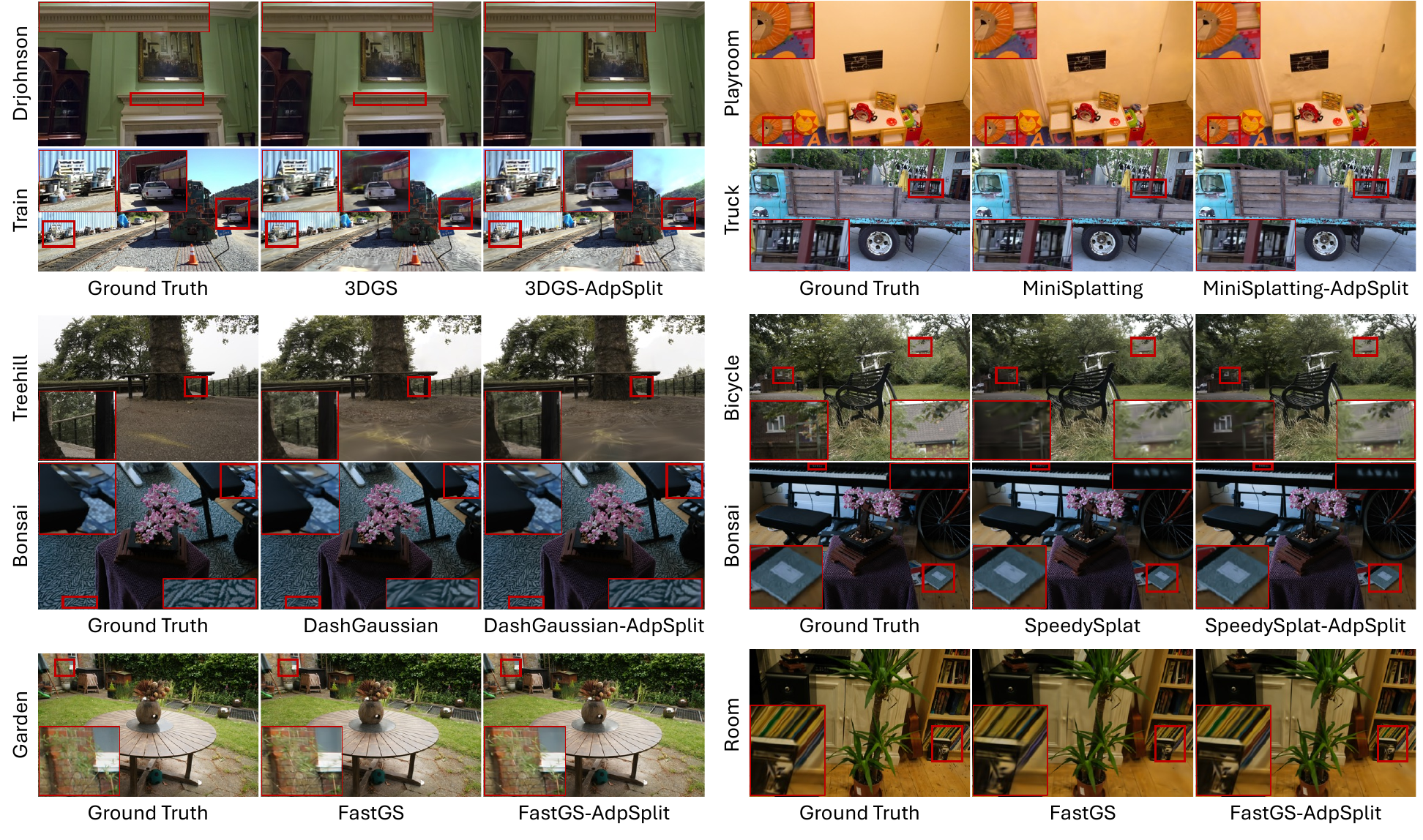}
    \caption{Qualitative results on the 
    scenes from MipNeRF360~\citep{Barron2022Mipnerf360}, Deep-Blending~\citep{Hedman2018Deep}, and Tanks\&Temples~\citep{Knapitsch2017Tanks}. 
    Our error-driven child initialization better reconstructs small details that fixed-cardinality random splitting fails to capture effectively, as highlighted by the red boxes.}
    \label{fig:5}
\end{figure}

The final number of Gaussians usually increases slightly relative to the
shortened schedule, but remains below the corresponding full densification
baseline in most cases. This suggests that AdpSplit does not simply add
uncontrolled Gaussians; instead, it allocates children to regions useful for
recovering scene structure. One exception is DashGaussian-AdpSplit, which can
take longer than DashGaussian-S despite using a comparable or smaller Gaussian
count. We attribute this to an overly deep cross-view child-merging graph formed
during low-resolution early training. Overall, these results support our claim
that error-driven split child generation can recover scene structure otherwise
missed when densification is shortened. Notably, our error-driven child initialization prioritizes the reconstruction of
small details. As shown in \autoref{fig:5}, AdpSplit deploys split children not
randomly, but in regions where they effectively reduce view loss, such as the
patterns on the book and carpet in the \texttt{Bonsai} scene and the railings in
the \texttt{Treehill} scene.

\subsubsection{Ablation study}

\begin{table}[t]
  \caption{Ablation results averaged over MipNeRF360 scenes. Time is reported in seconds and \#G. in millions.}
  \label{tab:2}
  \centering
  \resizebox{0.7\linewidth}{!}{%
  \begin{tabular}{@{}lrrrrrr@{}}
    \toprule
    \textbf{Method} & \textbf{PSNR} & \textbf{SSIM} & \textbf{LPIPS} & \textbf{Time} & \textbf{\#G.} & \textbf{FPS} \\
    \midrule
    3DGS & 27.53 & 0.813 & 0.221 & 1807 & 2.65 & 99.50 \\
    3DGS-S & 27.41 & 0.808 & 0.228 & 1318 & 2.41 & 109.98 \\
    \midrule
    + Adaptive $N$ & 27.47 & 0.810 & 0.224 & 1398 & 2.58 & 105.48 \\
    + Child initialization & 27.49 & 0.810 & 0.224 & 1436 & 2.63 & 102.44 \\
    + Child merging (=AdpSplit) & 27.54 & 0.810 & 0.225 & 1391 & 2.55 & 105.13 \\
    \bottomrule
  \end{tabular}
  }
\end{table}

\autoref{tab:2} ablates the components of AdpSplit on the
shortened 3DGS schedule. Randomly placing an adaptive number of children,
derived by counting partitioned error regions, recovers part of the lost
reconstruction quality, suggesting that a fixed $N=2$ split cannot
effectively search for useful scene structure. Child initialization further
improves PSNR, but adds more Gaussians to the scene. Child merging obtains the
highest PSNR, successfully recovers the PSNR of 3DGS, and removes some redundant
Gaussians while preserving the training-time reduction from the shortened
schedule.

\section{Conclusion}
We presented AdpSplit, an error-driven adaptive split operator for accelerating
3D Gaussian Splatting training. Instead of assigning each split candidate a
fixed number of randomly placed children, AdpSplit uses attributable
L1-pixel-error regions to initialize a dynamic number of children and their
parameters, then merges redundant cross-view proposals before insertion.
Because AdpSplit only replaces the split operator, it can be integrated into
existing 3DGS acceleration pipelines as a drop-in replacement.
Our experiments show that AdpSplit recovers much of the quality lost by
shortened densification schedules while retaining most of their speedup. Across
3DGS, MiniSplatting, DashGaussian, SpeedySplat, and FastGS on three benchmarks,
AdpSplit consistently maintains or improves the quality-efficiency trade-off
and reconstructs small details often missed by random child placement. These
results suggest that adaptive, error-driven child generation can reduce the
densification rounds needed for 3DGS scene discovery.

\paragraph{Limitations.}
AdpSplit is effective when densification is limited to 9k iterations, but more
elaborate splitting strategies may further reduce the required number of
densification steps. Since our error-driven child initialization starts from the
first densification round, overly large early Gaussians can cover broad scene
regions and may generate floating children from unreliable error regions. Future
work could add early-stage safeguards or extend adaptive splitting to dynamic
and sparse-view reconstruction settings.

\begin{ack}
Supported by the Intelligence Advanced Research Projects Activity (IARPA) via Department of Interior/ Interior Business Center (DOI/IBC) contract number 140D0423C0076. The U.S. Government is authorized to reproduce and distribute reprints for Governmental purposes notwithstanding any copyright annotation thereon. Disclaimer: The views and conclusions contained herein are those of the authors and should not be interpreted as necessarily representing the official policies or endorsements, either expressed or implied, of IARPA, DOI/IBC, or the U.S. Government.
\end{ack}

{
\small
\bibliographystyle{unsrtnat}
\bibliography{references}
}


\appendix

\section{Child Initialization Details}
\label{sec:child_position_derivation}

For each retained error region $\mathcal{R}_v^q$ of parent Gaussian
$\mathcal{G}_i$, we initialize one child Gaussian from the region's centroid,
covariance, and color. Let the camera intrinsics be
$(f_x,f_y,p_x,p_y)$, where $(f_x,f_y)$ are the focal lengths and
$(p_x,p_y)$ is the image center. The centroid
$(\bar{x}_v^q,\bar{y}_v^q)$ defines the camera-space ray direction
\begin{equation}
  \mathbf{d}_{\mathrm{cam},v}^q =
  \begin{bmatrix}
    (\bar{x}_v^q-p_x)/f_x \\
    (\bar{y}_v^q-p_y)/f_y \\
    1
  \end{bmatrix}.
\end{equation}
Given the camera-to-world rotation $\mathbf{R}_{c2w}$, the corresponding
unit ray direction in world coordinates is
\begin{equation}
  \hat{\mathbf{d}}_v^q =
  \frac{\mathbf{R}_{c2w}\mathbf{d}_{\mathrm{cam},v}^q}
       {\|\mathbf{R}_{c2w}\mathbf{d}_{\mathrm{cam},v}^q\|}.
\end{equation}
The back-projected camera ray is
\begin{equation}
  \boldsymbol{\ell}_v^q(t) = \mathbf{o}_v + t\hat{\mathbf{d}}_v^q,
\end{equation}
where $\mathbf{o}_v$ is the camera center in world coordinates.

We choose the point on this ray that minimizes the Mahalanobis distance to the
parent center $\boldsymbol{\mu}_i$:
\begin{equation}
  t_v^* = \arg\min_t
  \left(\boldsymbol{\ell}_v^q(t)-\boldsymbol{\mu}_i\right)^\top
  \boldsymbol{\Sigma}_i^{-1}
  \left(\boldsymbol{\ell}_v^q(t)-\boldsymbol{\mu}_i\right).
\end{equation}
Let $\mathbf{A}=\boldsymbol{\Sigma}_i^{-1}$ and
$\mathbf{b}=\boldsymbol{\mu}_i-\mathbf{o}_v$. Substituting the ray equation
gives
\begin{align}
  f(t)
  &= \left(t\hat{\mathbf{d}}_v^q-\mathbf{b}\right)^\top
     \mathbf{A}
     \left(t\hat{\mathbf{d}}_v^q-\mathbf{b}\right) \\
  &= t^2 (\hat{\mathbf{d}}_v^q)^\top\mathbf{A}\hat{\mathbf{d}}_v^q
     - 2t \mathbf{b}^\top\mathbf{A}\hat{\mathbf{d}}_v^q
     + \mathbf{b}^\top\mathbf{A}\mathbf{b}.
\end{align}
Taking the derivative with respect to $t$ and setting it to zero yields
\begin{equation}
  \frac{\partial f}{\partial t}
  =
  2t\,(\hat{\mathbf{d}}_v^q)^\top\mathbf{A}\hat{\mathbf{d}}_v^q
  - 2\mathbf{b}^\top\mathbf{A}\hat{\mathbf{d}}_v^q
  = 0.
\end{equation}
Therefore,
\begin{equation}
  t_v^*
  =
  \frac{\mathbf{b}^\top\mathbf{A}\hat{\mathbf{d}}_v^q}
       {(\hat{\mathbf{d}}_v^q)^\top\mathbf{A}\hat{\mathbf{d}}_v^q}
  =
  \frac{(\boldsymbol{\mu}_i - \mathbf{o}_v)^\top \boldsymbol{\Sigma}_i^{-1}
        \hat{\mathbf{d}}_v^q}
       {(\hat{\mathbf{d}}_v^q)^\top\boldsymbol{\Sigma}_i^{-1}\hat{\mathbf{d}}_v^q}.
\end{equation}
In implementation, we add a small $\epsilon$ to the denominator for numerical
stability, giving the expression in \autoref{eq:tstar}. The child center is
\begin{equation}
  \boldsymbol{\mu}_v^q = \boldsymbol{\ell}_v^q(t_v^*).
\end{equation}

Next, we convert the image-space region covariance into child scale and
orientation. Because $\mathbf{d}_{\mathrm{cam},v}^q$ is normalized when forming
$\hat{\mathbf{d}}_v^q$, the camera-space depth associated with $t_v^*$ is
\begin{equation}
  t_{z,v}^* = \frac{t_v^*}{\|\mathbf{d}_{\mathrm{cam},v}^q\|}.
\end{equation}
Let $\mathbf{u}_v$ and $\mathbf{w}_v$ be the camera right and down directions in
world coordinates, and let $s_{\max}^p=\max(\mathbf{s}_i)$ be the parent's
maximum scale. We clamp the unprojected axis components with
$\operatorname{clip}(x,a)=\operatorname{sign}(x)\min(|x|,a)$:
\begin{align}
  w_{1x,v}^q
  &= \operatorname{clip}
     \left(e_{1,x,v}^q\sigma_{1,v}^q\frac{t_{z,v}^*}{f_x},
           s_{\max}^p |e_{1,x,v}^q|\right), \\
  w_{1y,v}^q
  &= \operatorname{clip}
     \left(e_{1,y,v}^q\sigma_{1,v}^q\frac{t_{z,v}^*}{f_y},
           s_{\max}^p |e_{1,y,v}^q|\right), \\
  w_{2x,v}^q
  &= \operatorname{clip}
     \left(e_{2,x,v}^q\sigma_{2,v}^q\frac{t_{z,v}^*}{f_x},
           s_{\max}^p |e_{2,x,v}^q|\right), \\
  w_{2y,v}^q
  &= \operatorname{clip}
     \left(e_{2,y,v}^q\sigma_{2,v}^q\frac{t_{z,v}^*}{f_y},
           s_{\max}^p |e_{2,y,v}^q|\right).
\end{align}
The two PCA directions are then unprojected to world-space vectors:
\begin{align}
  \mathbf{a}_{1,v}^q &= w_{1x,v}^q\mathbf{u}_v + w_{1y,v}^q\mathbf{w}_v, \\
  \mathbf{a}_{2,v}^q &= w_{2x,v}^q\mathbf{u}_v + w_{2y,v}^q\mathbf{w}_v.
  \label{eq:axes_world}
\end{align}
The child scales are
\begin{equation}
  s_{1,v}^q = \|\mathbf{a}_{1,v}^q\|, \quad
  s_{2,v}^q = \|\mathbf{a}_{2,v}^q\|, \quad
  s_{3,v}^q = s_{2,v}^q.
  \label{eq:child_scales}
\end{equation}
Although the PCA directions are orthogonal in image space, their unprojected
vectors can lose exact orthogonality after perspective scaling and clamping. We
therefore obtain orthonormal in-plane axes by Gram-Schmidt before forming the
child rotation matrix:
\begin{equation}
  \hat{\mathbf{u}}_1 =
  \frac{\mathbf{a}_{1,v}^q}{\|\mathbf{a}_{1,v}^q\|}, \quad
  \hat{\mathbf{u}}_2 =
  \frac{
    \mathbf{a}_{2,v}^q
    - ((\mathbf{a}_{2,v}^q)^\top\hat{\mathbf{u}}_1)\hat{\mathbf{u}}_1
  }{
    \left\|
      \mathbf{a}_{2,v}^q
      - ((\mathbf{a}_{2,v}^q)^\top\hat{\mathbf{u}}_1)\hat{\mathbf{u}}_1
    \right\|
  }.
  \label{eq:gram_schmidt}
\end{equation}
With camera forward direction $\mathbf{z}_v$, the child covariance is
\begin{equation}
  \mathbf{R}_v^q =
  \bigl[\hat{\mathbf{u}}_1 \mid \hat{\mathbf{u}}_2 \mid \mathbf{z}_v\bigr],
  \quad
  \boldsymbol{\Sigma}_v^q =
  \mathbf{R}_v^q
  \operatorname{diag}((s_{1,v}^q)^2,(s_{2,v}^q)^2,(s_{3,v}^q)^2)
  {\mathbf{R}_v^q}^\top.
  \label{eq:child_cov}
\end{equation}

Finally, we initialize the child color from the ground-truth RGB value
$\mathbf{r}_v^q$ at the region centroid, convert it to the DC spherical-harmonic
coefficient, and set all higher-order SH coefficients to zero. The child opacity
inherits the parent opacity, matching the vanilla split initialization.

\section{Additional results}
\label{sec:additional_results}

\begin{table}[t]
  \caption{Average peak GPU memory (GB) on MipNeRF360, Deep-Blending, and Tanks\&Temples.}
  \label{tab:3}
  \centering
  \begin{tabular}{@{}lrrr@{}}
    \toprule
    \textbf{Method} & \textbf{MipNeRF360} & \textbf{Deep-Blending} & \textbf{Tanks\&Temples} \\
    \midrule
    3DGS & 9.93 & 8.14 & 4.75 \\
    3DGS-S & \textbf{9.52} & \textbf{7.57} & \textbf{4.55} \\
    3DGS-AdpSplit & 10.00 & 7.80 & 4.77 \\
    \midrule
    MiniSplatting & 6.33 & 6.20 & 4.65 \\
    MiniSplatting-S & \textbf{6.32} & \textbf{6.07} & \textbf{4.56} \\
    MiniSplatting-AdpSplit & 6.57 & 6.33 & 4.83 \\
    \midrule
    DashGaussian & 8.67 & 7.75 & 4.49 \\
    DashGaussian-S & \textbf{8.61} & 7.60 & 4.44 \\
    DashGaussian-AdpSplit & 8.65 & \textbf{7.51} & \textbf{4.36} \\
    \midrule
    SpeedySplat & 9.01 & 6.85 & \textbf{4.50} \\
    SpeedySplat-S & \textbf{8.87} & \textbf{6.83} & \textbf{4.50} \\
    SpeedySplat-AdpSplit & 9.50 & 7.11 & 4.74 \\
    \midrule
    FastGS & 7.06 & 4.98 & 2.71 \\
    FastGS-S & \textbf{5.08} & \textbf{3.69} & \textbf{2.22} \\
    FastGS-AdpSplit & 5.83 & 3.94 & 2.39 \\
    \bottomrule
  \end{tabular}
\end{table}

\begin{figure}[t]
    \centering
    \includegraphics[width=\linewidth]{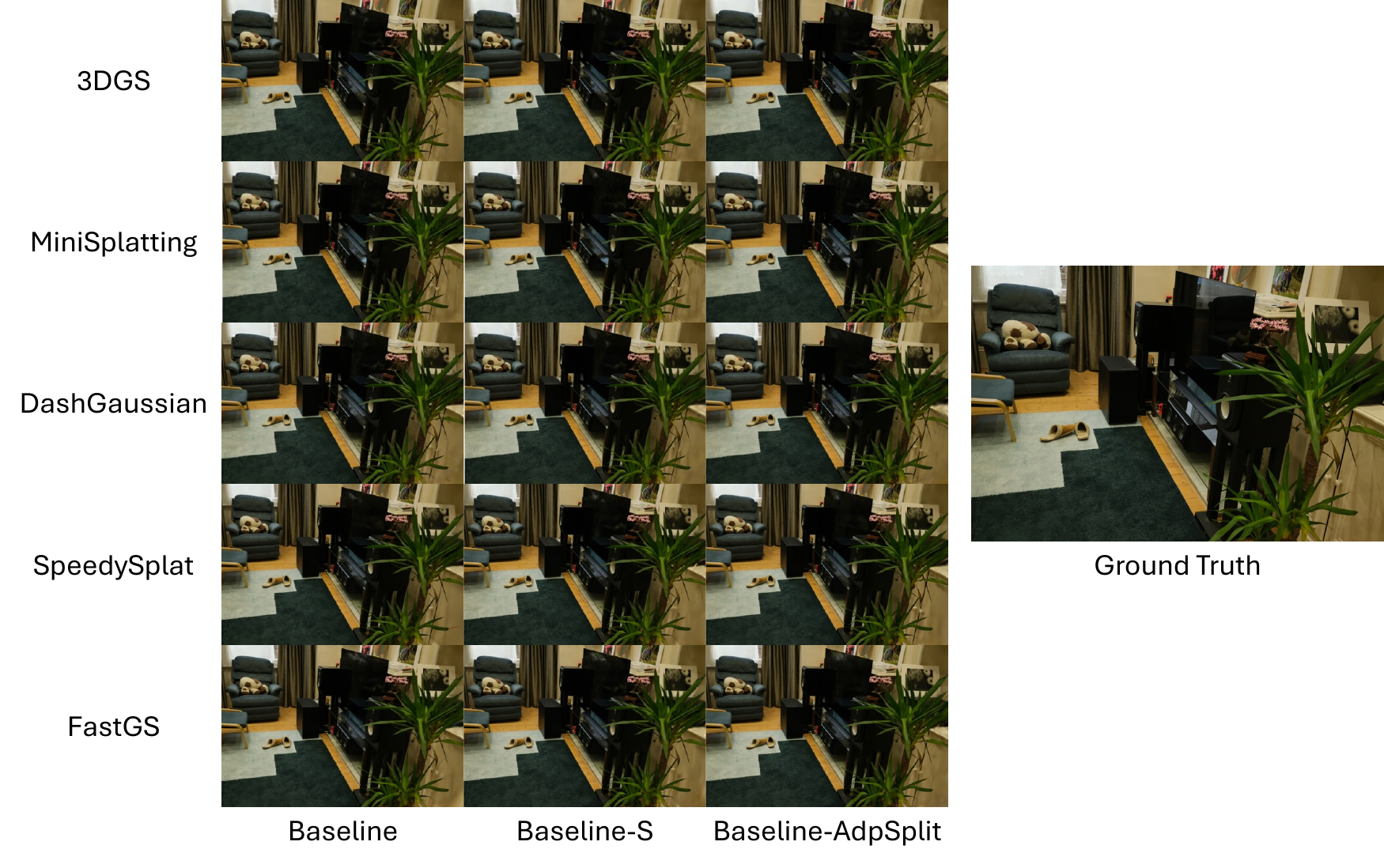}
    \caption{Qualitative results on the \texttt{Room} scene from MipNeRF360}
    \label{fig:6}
\end{figure}
\begin{figure}[t]
    \centering
    \includegraphics[width=\linewidth]{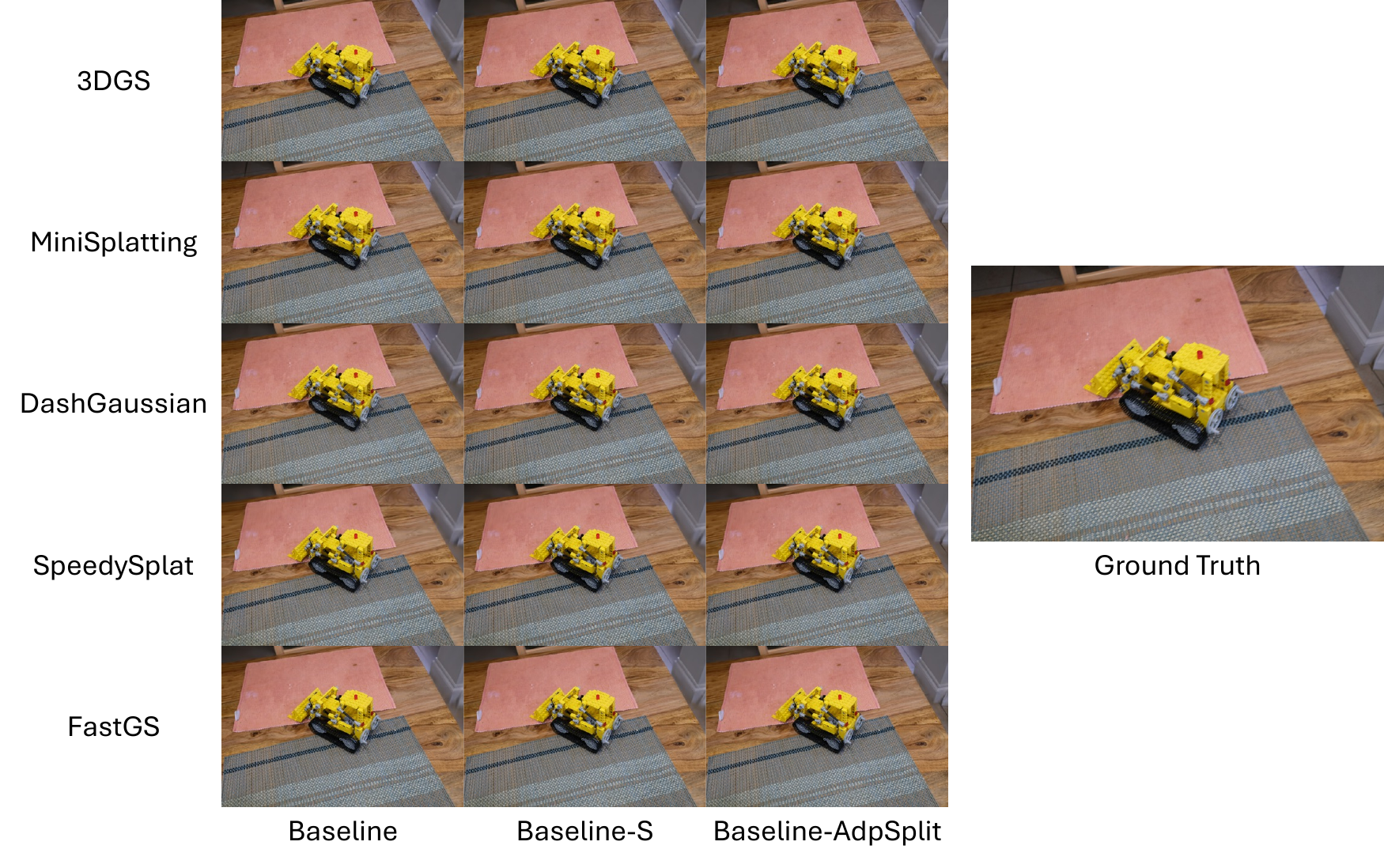}
    \caption{Qualitative results on the \texttt{Kitchen} scene from MipNeRF360}
    \label{fig:7}
\end{figure}
\begin{figure}[t]
    \centering
    \includegraphics[width=\linewidth]{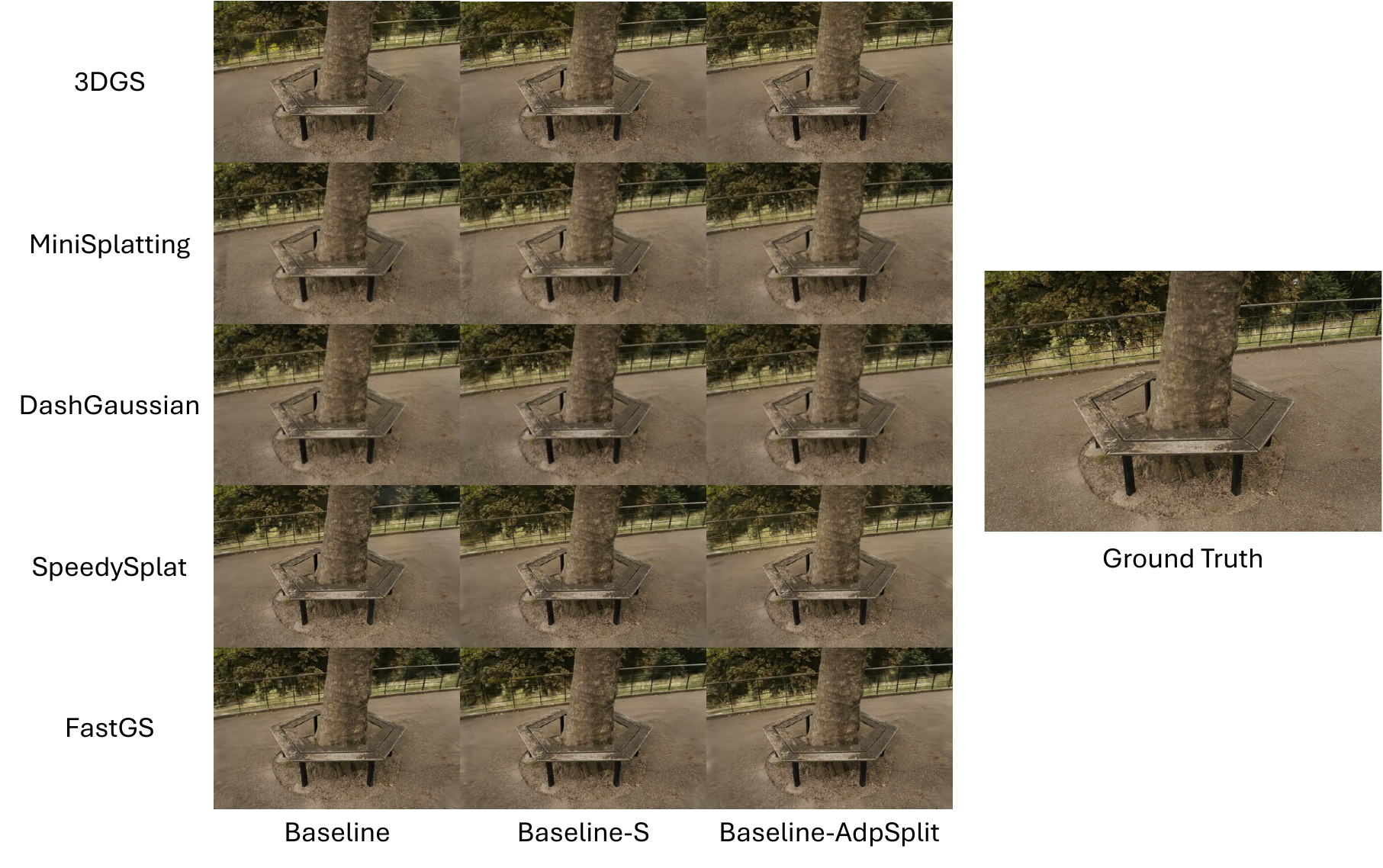}
    \caption{Qualitative results on the \texttt{Treehill} scene from MipNeRF360}
    \label{fig:8}
\end{figure}
\begin{figure}[t]
    \centering
    \includegraphics[width=\linewidth]{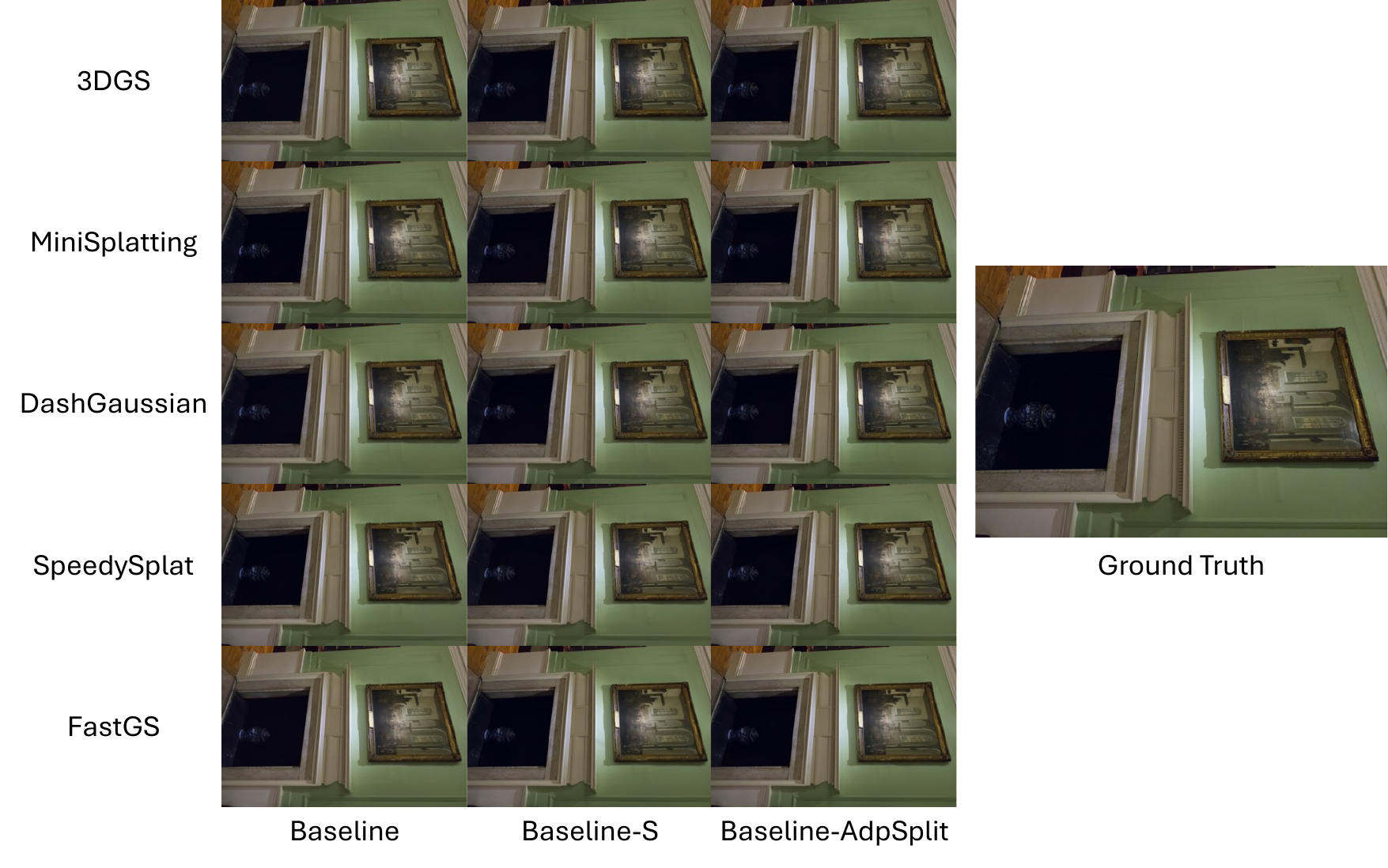}
    \caption{Qualitative results on the \texttt{Drjohnson} scene from Deep-Blending}
    \label{fig:9}
\end{figure}
\begin{figure}[t]
    \centering
    \includegraphics[width=\linewidth]{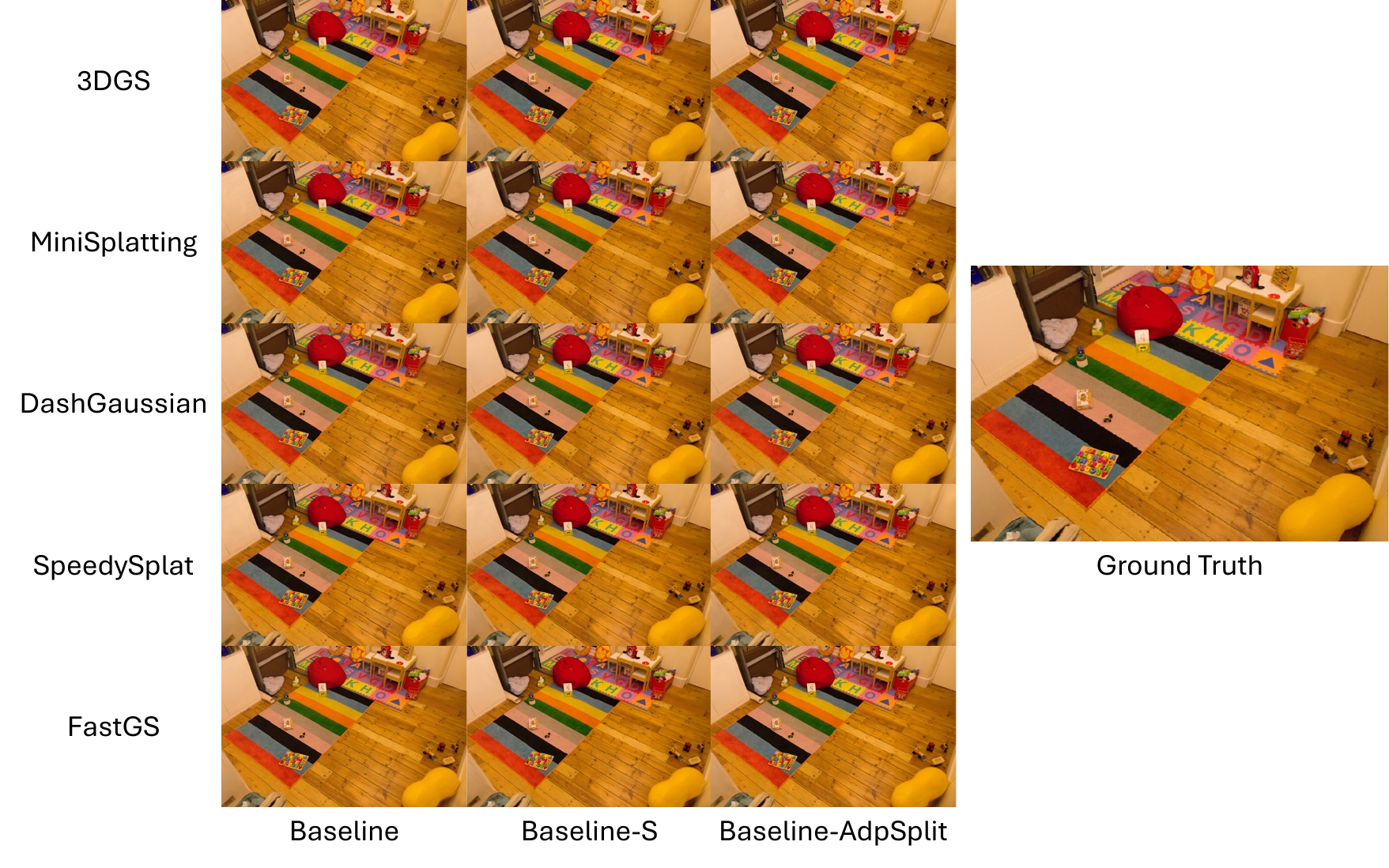}
    \caption{Qualitative results on the \texttt{Playroom} scene from Deep-Blending}
    \label{fig:10}
\end{figure}
\begin{figure}[t]
    \centering
    \includegraphics[width=\linewidth]{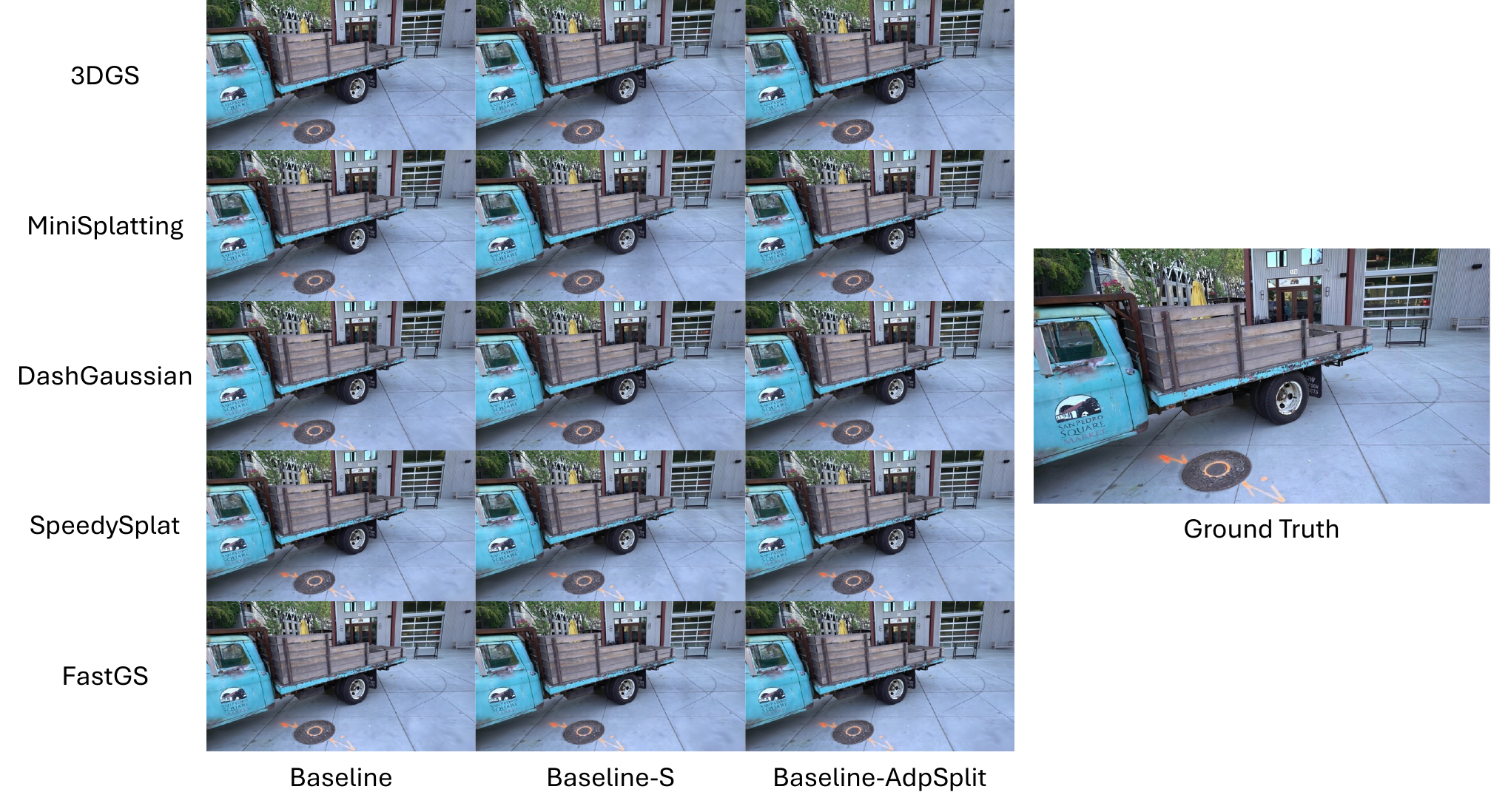}
    \caption{Qualitative results on the \texttt{Truck} scene from Tanks\&Temples}
    \label{fig:11}
\end{figure}

\autoref{tab:3} reports the peak GPU memory values corresponding to the main
results in \autoref{tab:1}. \autoref{fig:6}, \autoref{fig:7}, \autoref{fig:8},
\autoref{fig:9}, \autoref{fig:10}, and \autoref{fig:11} present additional
visualizations. The first column shows the baseline method, the second column
shows the baseline with a shortened densification schedule, and the last column
shows the corresponding AdpSplit variant.

\clearpage
\subsection{Hyperparameter Sensitivity}
\label{ssec:hyperparameter}

\begin{table}[t]
  \caption{Tuning results for erosion length $r_{\mathrm{erode}}$, averaged over nine MipNeRF360 scenes.}
  \label{tab:adpsplit_error_erode_tuning}
  \centering
  \begin{tabular}{@{}lrrrrrrr@{}}
    \toprule
    \textbf{$r_{\mathrm{erode}}$} & \textbf{PSNR} & \textbf{SSIM} & \textbf{LPIPS} & \textbf{Time} & \textbf{\#G.} & \textbf{FPS} & \textbf{PM} \\
    \midrule
    0 & 27.48 & 0.811 & 0.221 & 1100 & 2.69 & 126.17 & 16.88 \\
    1 & 27.47 & 0.810 & 0.224 & 1109 & 2.63 & 125.06 & 11.58 \\
    2 & 27.54 & 0.810 & 0.225 & 1107 & 2.55 & 125.36 & 10.00 \\
    3 & 27.53 & 0.810 & 0.226 & 1110 & 2.49 & 125.89 & 9.92 \\
    \bottomrule
  \end{tabular}
\end{table}

\begin{table}[t]
  \caption{Tuning results for the number of error-level bands $L$, averaged over nine MipNeRF360 scenes.}
  \label{tab:adpsplit_l_levels_tuning}
  \centering
  \begin{tabular}{@{}lrrrrrrr@{}}
    \toprule
    \textbf{$L$} & \textbf{PSNR} & \textbf{SSIM} & \textbf{LPIPS} & \textbf{Time} & \textbf{\#G.} & \textbf{FPS} & \textbf{PM} \\
    \midrule
    1 & 27.46 & 0.810 & 0.225 & 1111 & 2.53 & 125.39 & 9.99 \\
    2 & 27.52 & 0.810 & 0.225 & 1107 & 2.55 & 125.49 & 10.00 \\
    3 & 27.54 & 0.810 & 0.225 & 1107 & 2.55 & 125.36 & 10.00 \\
    4 & 27.49 & 0.810 & 0.225 & 1107 & 2.56 & 125.17 & 10.02 \\
    \bottomrule
  \end{tabular}
\end{table}

\begin{table}[t]
  \caption{Tuning results for the child cap $N_{\max}$, averaged separately over outdoor and indoor MipNeRF360 scenes.}
  \label{tab:adpsplit_max_child_n_tuning}
  \centering
  \setlength{\tabcolsep}{2pt}
  \resizebox{\linewidth}{!}{%
  \begin{tabular}{@{}lrrrrrrr rrrrrrr@{}}
    \toprule
    \textbf{$N_{\max}$} & \multicolumn{7}{c}{\textbf{Outdoor}} & \multicolumn{7}{c}{\textbf{Indoor}} \\
    \cmidrule(lr){2-8} \cmidrule(lr){9-15}
    & \textbf{PSNR} & \textbf{SSIM} & \textbf{LPIPS} & \textbf{Time} & \textbf{\#G.} & \textbf{FPS} & \textbf{PM}
    & \textbf{PSNR} & \textbf{SSIM} & \textbf{LPIPS} & \textbf{Time} & \textbf{\#G.} & \textbf{FPS} & \textbf{PM} \\
    \midrule
    4 & 24.58 & 0.714 & 0.271 & 1139 & 3.52 & 110.77 & 9.95 & 31.17 & 0.931 & 0.168 & 1062 & 1.26 & 145.48 & 9.96 \\
    9 & 24.58 & 0.714 & 0.270 & 1143 & 3.57 & 109.72 & 10.02 & 31.23 & 0.931 & 0.168 & 1062 & 1.26 & 145.41 & 9.97 \\
    19 & 24.53 & 0.714 & 0.270 & 1148 & 3.62 & 110.20 & 10.09 & 31.23 & 0.931 & 0.168 & 1062 & 1.26 & 144.90 & 9.97 \\
    29 & 24.56 & 0.713 & 0.270 & 1152 & 3.63 & 109.28 & 10.11 & 31.19 & 0.930 & 0.168 & 1062 & 1.26 & 145.22 & 9.98 \\
    all & 24.54 & 0.713 & 0.270 & 1159 & 3.70 & 108.28 & 10.21 & 31.22 & 0.930 & 0.168 & 1067 & 1.27 & 144.08 & 9.98 \\
    \bottomrule
  \end{tabular}%
  }
\end{table}

\begin{table}[t]
  \caption{Tuning results for minimum error-region area $m_{\min}$, averaged over nine MipNeRF360 scenes.}
  \label{tab:adpsplit_minimum_pixels_tuning}
  \centering
  \begin{tabular}{@{}lrrrrrrr@{}}
    \toprule
    \textbf{$m_{\min}$} & \textbf{PSNR} & \textbf{SSIM} & \textbf{LPIPS} & \textbf{Time} & \textbf{\#G.} & \textbf{FPS} & \textbf{PM} \\
    \midrule
    0 & 27.52 & 0.810 & 0.224 & 1114 & 2.62 & 124.76 & 11.73 \\
    5 & 27.54 & 0.810 & 0.225 & 1107 & 2.55 & 125.36 & 10.00 \\
    10 & 27.50 & 0.810 & 0.225 & 1107 & 2.52 & 125.77 & 9.96 \\
    \bottomrule
  \end{tabular}
\end{table}

\begin{table}[t]
  \caption{Tuning results for the number of sampled views $V$, averaged separately over outdoor and indoor MipNeRF360 scenes.}
  \label{tab:adpsplit_num_cams_tuning}
  \centering
  \setlength{\tabcolsep}{2pt}
  \resizebox{\linewidth}{!}{%
  \begin{tabular}{@{}lrrrrrrr rrrrrrr@{}}
    \toprule
    \textbf{$V$} & \multicolumn{7}{c}{\textbf{Outdoor}} & \multicolumn{7}{c}{\textbf{Indoor}} \\
    \cmidrule(lr){2-8} \cmidrule(lr){9-15}
    & \textbf{PSNR} & \textbf{SSIM} & \textbf{LPIPS} & \textbf{Time} & \textbf{\#G.} & \textbf{FPS} & \textbf{PM}
    & \textbf{PSNR} & \textbf{SSIM} & \textbf{LPIPS} & \textbf{Time} & \textbf{\#G.} & \textbf{FPS} & \textbf{PM} \\
    \midrule
    5 & 24.57 & 0.713 & 0.272 & 1129 & 3.46 & 112.09 & 10.78 & 31.16 & 0.931 & 0.169 & 1034 & 1.23 & 148.06 & 9.62 \\
    10 & 24.58 & 0.714 & 0.270 & 1143 & 3.57 & 109.72 & 10.02 & 31.19 & 0.931 & 0.168 & 1059 & 1.25 & 145.31 & 9.75 \\
    20 & 24.58 & 0.714 & 0.269 & 1157 & 3.64 & 108.96 & 10.99 & 31.23 & 0.931 & 0.168 & 1062 & 1.26 & 144.90 & 9.97 \\
    30 & 24.57 & 0.714 & 0.269 & 1177 & 3.70 & 108.22 & 22.30 & 31.27 & 0.931 & 0.167 & 1073 & 1.27 & 144.40 & 10.18 \\
    50 & 24.60 & 0.714 & 0.268 & 1196 & 3.80 & 106.55 & 29.35 & 31.23 & 0.931 & 0.167 & 1097 & 1.30 & 142.15 & 11.05 \\
    \bottomrule
  \end{tabular}%
  }
\end{table}

\autoref{tab:adpsplit_error_erode_tuning}--\autoref{tab:adpsplit_num_cams_tuning}
summarize the average effect of AdpSplit hyper-parameters in the updated tuning
sheet. The first three tables average over all nine MipNeRF360 scenes. For
$N_{\max}$ and $V$, we separately report outdoor and indoor averages. Time is
reported in seconds, \#G. in millions, and PM
denotes peak GPU memory in GB. All experiments in this section were run on a
single NVIDIA A100 80GB GPU.


\clearpage
\input{checklist.tex}

\end{document}

%% file: checklist.tex
\section*{NeurIPS Paper Checklist}

\begin{enumerate}

\item {\bf Claims}
    \item[] Question: Do the main claims made in the abstract and introduction accurately reflect the paper's contributions and scope?
    \item[] Answer: \answerYes{} 
    \item[] Justification: AdpSplit is able to recover scene reconstruction quality lost due to a shortened densification schedule by adaptively adding new Gaussians in high-error regions that indicate under-reconstruction, as demonstrated in Table 1.

\item {\bf Limitations}
    \item[] Question: Does the paper discuss the limitations of the work performed by the authors?
    \item[] Answer: \answerYes{} 
    \item[] Justification: We discuss the limitations in Conclusion section.

\item {\bf Theory assumptions and proofs}
    \item[] Question: For each theoretical result, does the paper provide the full set of assumptions and a complete (and correct) proof?
    \item[] Answer: \answerNA{} 
    \item[] Justification: The paper doesn't include theoretical results.

    \item {\bf Experimental result reproducibility}
    \item[] Question: Does the paper fully disclose all the information needed to reproduce the main experimental results of the paper to the extent that it affects the main claims and/or conclusions of the paper (regardless of whether the code and data are provided or not)?
    \item[] Answer: \answerYes{} 
    \item[] Justification: The paper provides the results needed and ablation study in Figure 2 and section 5. More results are listed in Appendix B.

\item {\bf Open access to data and code}
    \item[] Question: Does the paper provide open access to the data and code, with sufficient instructions to faithfully reproduce the main experimental results, as described in supplemental material?
    \item[] Answer: \answerNo{} 
    \item[] Justification: The datasets are all publicly accessible in the internet. We don't provide our code at the time of submission. We will open it when paper is accepted.

\item {\bf Experimental setting/details}
    \item[] Question: Does the paper specify all the training and test details (e.g., data splits, hyperparameters, how they were chosen, type of optimizer) necessary to understand the results?
    \item[] Answer: \answerYes{} 
    \item[] Justification: The detailed experiment settings are listed in Section 5.1.

\item {\bf Experiment statistical significance}
    \item[] Question: Does the paper report error bars suitably and correctly defined or other appropriate information about the statistical significance of the experiments?
    \item[] Answer: \answerNA{} 
    \item[] Justification: This paper follows the existing work about 3DGS and lists the essential detailed quantitative results in Section 5.2 and Appendix B. But it does not list error bars.

\item {\bf Experiments compute resources}
    \item[] Question: For each experiment, does the paper provide sufficient information on the computer resources (type of compute workers, memory, time of execution) needed to reproduce the experiments?
    \item[] Answer: \answerYes{} 
    \item[] Justification: This paper lists the hardware used in Section 5.2 and Appendix B. This paper shows the graphic memory required in Table 2 and Appendix B.
    
\item {\bf Code of ethics}
    \item[] Question: Does the research conducted in the paper conform, in every respect, with the NeurIPS Code of Ethics \url{https://neurips.cc/public/EthicsGuidelines}?
    \item[] Answer: \answerYes{} 
    \item[] Justification: The research conducted in this paper confirm the NeurIPS Code of Conducts.

\item {\bf Broader impacts}
    \item[] Question: Does the paper discuss both potential positive societal impacts and negative societal impacts of the work performed?
    \item[] Answer: \answerNA{} 
    \item[] Justification: This paper primarily focuses on improving model training efficiency. It is not expected to have any societal impact.
    
\item {\bf Safeguards}
    \item[] Question: Does the paper describe safeguards that have been put in place for responsible release of data or models that have a high risk for misuse (e.g., pre-trained language models, image generators, or scraped datasets)?
    \item[] Answer: \answerNA{} 
    \item[] Justification: This paper does not pose such risk.

\item {\bf Licenses for existing assets}
    \item[] Question: Are the creators or original owners of assets (e.g., code, data, models), used in the paper, properly credited and are the license and terms of use explicitly mentioned and properly respected?
    \item[] Answer: \answerYes{} 
    \item[] Justification: This paper follows the applicable licenses and terms of usage.

\item {\bf New assets}
    \item[] Question: Are new assets introduced in the paper well documented and is the documentation provided alongside the assets?
    \item[] Answer: \answerNo{} 
    \item[] Justification: This paper provides implementation codes as new assets. The code and documentation will be open after paper is accepted.

\item {\bf Crowdsourcing and research with human subjects}
    \item[] Question: For crowdsourcing experiments and research with human subjects, does the paper include the full text of instructions given to participants and screenshots, if applicable, as well as details about compensation (if any)? 
    \item[] Answer: \answerNA{} 
    \item[] Justification: This paper does not involve crowdsourcing nor research with human subjects.

\item {\bf Institutional review board (IRB) approvals or equivalent for research with human subjects}
    \item[] Question: Does the paper describe potential risks incurred by study participants, whether such risks were disclosed to the subjects, and whether Institutional Review Board (IRB) approvals (or an equivalent approval/review based on the requirements of your country or institution) were obtained?
    \item[] Answer: \answerNA{} 
    \item[] Justification: This paper does not involve crowdsourcing nor research with human subjects.

\item {\bf Declaration of LLM usage}
    \item[] Question: Does the paper describe the usage of LLMs if it is an important, original, or non-standard component of the core methods in this research? Note that if the LLM is used only for writing, editing, or formatting purposes and does \emph{not} impact the core methodology, scientific rigor, or originality of the research, declaration is not required.
    \item[] Answer: \answerNA{} 
    \item[] Justification: LLMs are used solely for grammar improvement.

\end{enumerate}